\documentclass[pdflatex,sn-mathphys-num]{sn-jnl}% Math and Physical Sciences Numbered Reference Style
\usepackage{graphicx}%
\usepackage{multirow}%
\usepackage{amsmath,amssymb,amsfonts}%
\usepackage{amsthm}%
\usepackage{mathrsfs}%
\usepackage[title]{appendix}%
\usepackage{xcolor}%
\usepackage{textcomp}%
\usepackage{manyfoot}%
\usepackage{booktabs}%
\usepackage{algorithm}%
\usepackage{algorithmicx}%
\usepackage{algpseudocode}%
\usepackage{listings}%
\usepackage{array}%
\usepackage{bbding}%
\theoremstyle{thmstyleone}%
\theoremstyle{thmstyletwo}%

\theoremstyle{thmstylethree}%

\begin{document}

\title[Multi-Modal Artificial Intelligence of Embryo Grading and Pregnancy Prediction in Assisted Reproductive Technology: A Review]{Multi-Modal Artificial Intelligence of Embryo Grading and Pregnancy Prediction in Assisted Reproductive Technology: A Review}


\author[1,2]{\fnm{Xueqiang} \sur{Ouyang}}

\author*[2]{\fnm{Jia} \sur{Wei}}\email{csjwei@scut.edu.cn}

\affil[1]{\orgdiv{Information Department}, \orgname{The People's Hospital of Baoan Shenzhen}, \orgaddress{\city{Shenzhen}, \postcode{518101}, \state{Guangdong}, \country{China}}}
\affil[2]{\orgdiv{School of Computer Science and Engineering}, \orgname{South China University of Technology}, \orgaddress{\city{Guangzhou}, \postcode{510006}, \state{Guangdong}, \country{China}}}


\abstract{Infertility, a pressing global health concern, affects a substantial proportion of individuals worldwide.  While advancements in assisted reproductive technology (ART) have offered effective interventions, conventional in vitro fertilization-embryo transfer (IVF-ET) procedures still encounter significant hurdles in enhancing pregnancy success rates.  Key challenges include the inherent subjectivity in embryo grading and the inefficiency of multi-modal data integration.  Against this backdrop, the adoption of AI-driven technologies has emerged as a pivotal strategy to address these issues.  This article presents a comprehensive review of the progress in AI applications for embryo grading and pregnancy prediction from a novel perspective, with a specific focus on the utilization of different modal data, such as static images, time-lapse videos, and structured tabular data. The reason for this perspective is that reorganizing tasks based on data sources can not only more accurately depict the essence of the problem but also help clarify the rationality and limitations of model design. Furthermore, this review critically examines the core challenges in contemporary research, encompassing the intricacies of multi-modal feature fusion, constraints imposed by data scarcity, limitations in model generalization capabilities, and the dynamically evolving legal and regulatory frameworks. On this basis, it explicitly identifies potential avenues for future research, aiming to provide actionable guidance for advancing the application of multi-modal AI in the field of ART.}




\maketitle

\section{Introduction}\label{sec1}
Study has shown that up to 10$\%$-15$\%$ of couples of childbearing age are diagnosed with infertility worldwide \cite{1}. And as women age, especially over the age of 35, their fertility declines and the risk of pregnancy increases \cite{agepregnancy}. Infertility has long been a significant challenge for both families and society. Therefore, helping infertile couples conceive through modern medical technology is of great significance for both families and society. For families, prolonged infertility often places considerable psychological pressure on both partners \cite{infertilitystress}. The emergence of Assisted Reproductive Technology (ART) offers new hope by improving psychological well-being and overall family happiness. On a societal level, the adoption of ART provides more reproductive choices for families in different regions \cite{fertilitychoice}.\\
\indent ART consists of multiple therapeutic approaches designed to address infertility. Among these, In Vitro Fertilization-Embryo Transfer (IVF-ET) has emerged as the main and most established intervention, both in clinical practice and research \cite{2}. In this technique, oocyte and sperm are fertilized outside the body to form an embryo, which is then transferred back to the woman's uterus to achieve pregnancy \cite{111}. According to clinical statistics, the pregnancy success rate of a single transfer cycle of IVF-ET is only 40$\%$-50$\%$, and the final live birth rate is as low as 30$\%$ \cite{3}. There are many reasons for this result. First, the selection of high-quality embryos for uterine transfer primarily relies on embryologists’ visual assessment of embryo images, where higher-grade embryos are prioritized. However, this method is highly subjective and error prone, because the results of multiple assessments by the same embryologist or the same assessment by different embryologists are quite different \cite{4}. Secondly, the key factors affecting pregnancy include not only embryo quality, but also parental fertility indicators, such as parental age, number of high-quality embryos, endometrial thickness, sperm quality and other clinical indicators \cite{5}. Reproductive doctors need to predict the success of pregnancy according to embryo quality and fertility indicators of the couple, which once again introduces subjectivity and uncertainty. Moreover, reproductive doctors and embryologists are not the same group of people, so the coordination between different departments once again affects the outcome of pregnancy. As a result, the low success rate associated with traditional clinical decision-making in IVF-ET often necessitates multiple embryo transfers to achieve a successful pregnancy.  This leads to prolonged treatment timelines and higher financial burdens \cite{cost}. Moreover, patients who have experienced failed IVF-ET cycles are often reluctant to continue treatment, primarily due to the complexity and time-intensiveness of ART, which imposes significant psychological burdens on them \cite{6}.\\
\begin{figure}[t]
  % \centering
  \includegraphics[width=0.9\textwidth]{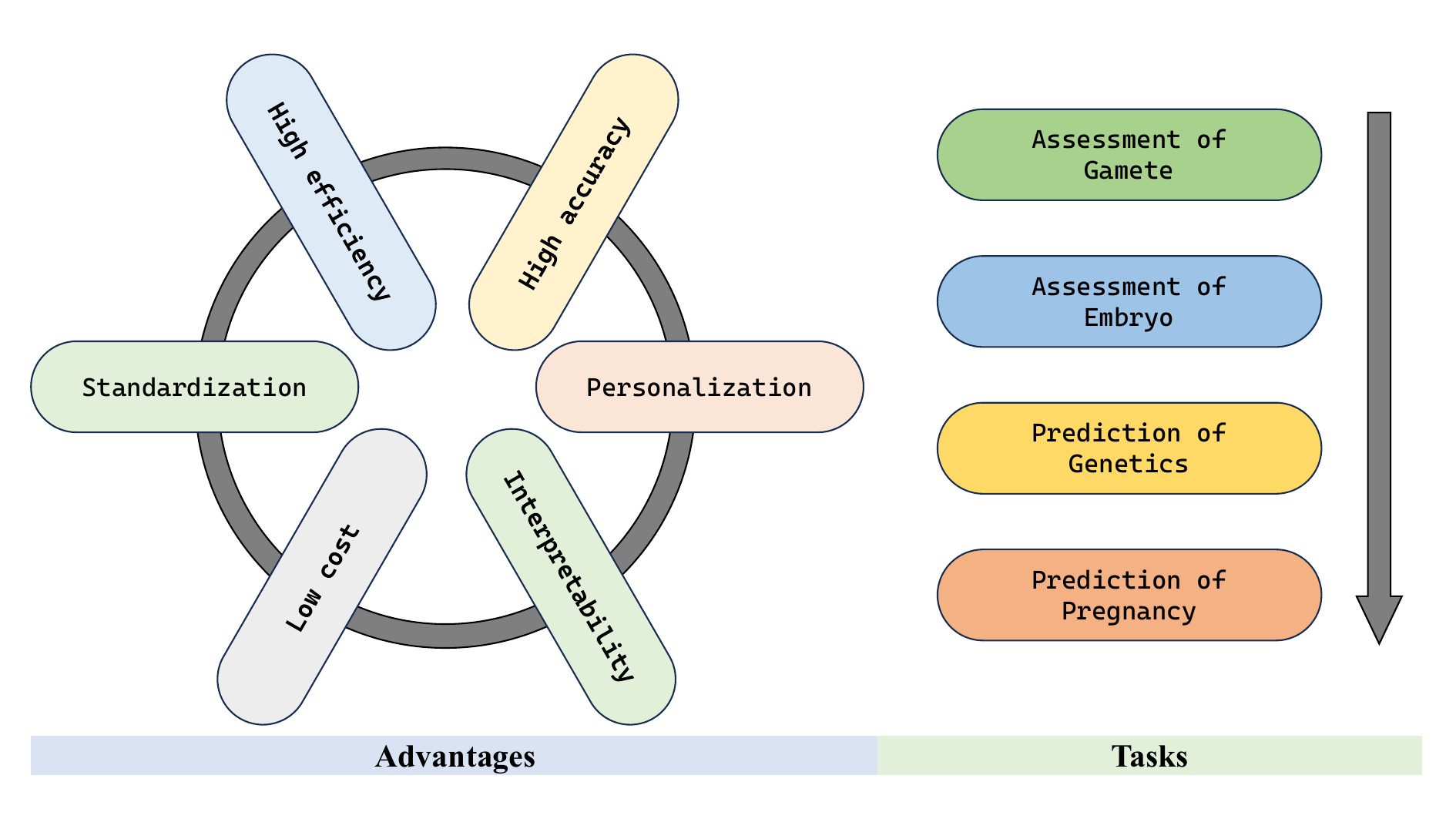}
  \caption{The advantages and tasks of AI in ART. The tasks are the AI technology tasks that may be applied at each stage of the IVF-ET process.}
  \label{adventage}
\end{figure}
\indent The aforementioned challenges underscore the urgent need for automated AI-based embryo grading and pregnancy prediction systems in clinical practice. These systems deliver transformative benefits across six key dimensions. (1) Standardization: Automated systems standardize embryo quality assessment through robust algorithms, eliminating subjective bias and ensuring consistent grading across operators and settings. (2) High efficiency: AI enables rapid processing of large-scale embryo data, significantly reducing manual workload and allowing clinicians to focus on complex cases and patient interactions. (3) High accuracy: By minimizing human error, automated grading enhances the reliability of embryo grading and pregnancy prediction, improving diagnostic precision. (4) Low cost: Automation reduces reliance on repeated manual assessments, lowering operational costs and increasing the accessibility and sustainability of advanced reproductive care. (5) Personalization: AI integrates multi-modal patient data to generate tailored treatment recommendations, supporting individualized clinical decision-making and care pathways. (6) Interpretability: AI models provide interpretable insights, revealing key factors behind predictions and fostering transparency and trust in clinical applications. The main advantages of AI in ART are shown in Fig. \ref{adventage}.\\
\indent Compared to traditional methods for embryo grading and pregnancy prediction, AI-driven approaches have demonstrated substantial advantages, leading to their widespread adoption in the field of ART \cite{aiinart}. While existing studies have explored various data modalities and addressed a range of specific tasks, a systematic review that comprehensively synthesizes these findings from the perspective of underlying data modalities remains absent. To address this gap, the present review adopts a novel focus on underlying data modalities, systematically summarizing and analyzing the current state of research on AI applications in embryo grading and pregnancy prediction. This helps researchers gain a comprehensive understanding of multi-modal AI in the field of assisted reproduction.\\
\indent Furthermore, this review identifies key future research directions, including: (1) advanced multi-modal feature fusion techniques to integrate diverse data sources for improved predictive performance; (2) innovative strategies to address data scarcity, such as synthetic data generation; (3) the enhancement of model generalization through cross-center validation and robust algorithm design; and (4) the development of ethical frameworks to ensure transparency, fairness, and compliance with evolving regulations. Addressing these challenges is critical, as they significantly impact the accuracy, generalizability, and usability of AI-driven embryo grading and pregnancy prediction in ART. By providing a clear theoretical foundation and contextual understanding, this review aims to facilitate future advancements in the field.
\begin{table*}[htbp]
  \centering
  \caption{A comparative analysis of current AI in ART reviews.}
  \resizebox{0.95\textwidth}{!}{
    \begin{tabular}{l|c|c|c|c|c|c}
    \hline
    \hline
    Papers& Gamete Assessment & Embryo Assessment & Gene Prediction &Pregnancy Prediction&Robot Automation & Multi-Modal \\
    \hline
     \cite{7}&  \Checkmark & \XSolidBrush &  \XSolidBrush & \XSolidBrush& \XSolidBrush& \XSolidBrush\\
    \hline
    \cite{8}\cite{blastocystreview}&  \XSolidBrush& \Checkmark &  \XSolidBrush &\XSolidBrush & \XSolidBrush& \XSolidBrush\\
    \hline
    \cite{9}\cite{10}&  \Checkmark& \Checkmark &  \XSolidBrush &\XSolidBrush & \XSolidBrush& \XSolidBrush\\
    \hline
    \cite{14}\cite{11}\cite{12}&  \XSolidBrush & \Checkmark &  \XSolidBrush &\Checkmark& \XSolidBrush& \XSolidBrush\\
    \hline
    \cite{17}&  \Checkmark& \Checkmark &  \Checkmark &\Checkmark  &\Checkmark & \XSolidBrush\\
    \hline
    \cite{18}\cite{13}&  \Checkmark& \Checkmark &  \Checkmark  &\Checkmark  & \XSolidBrush& \XSolidBrush\\
    \hline
    \cite{19}&  \Checkmark& \Checkmark &  \Checkmark &\XSolidBrush & \XSolidBrush& \XSolidBrush\\
    \hline
    \cite{20}& \XSolidBrush & \Checkmark &  \Checkmark  &\XSolidBrush & \XSolidBrush& \XSolidBrush\\
    \hline
    \cite{assessment}&  \Checkmark & \Checkmark & \XSolidBrush  &\XSolidBrush & \Checkmark & \XSolidBrush\\
    \hline
    \cite{15}\cite{16}&  \XSolidBrush& \XSolidBrush&  \XSolidBrush &\XSolidBrush & \Checkmark& \XSolidBrush\\
    \hline
    Ours& \XSolidBrush& \Checkmark &  \XSolidBrush &\Checkmark& \XSolidBrush& \Checkmark\\
    \hline
    \end{tabular}%
  \label{tablereview}%
  }
\end{table*}%
\section{Related Work}\label{sec2}
Dalal et al. \cite{7} systematically introduced the tasks of gamete assessment for sperm and oocytes, and summarized a variety of methods for automated analysis, which laid a foundation for the objective and standardized assessment of gamete quality. Furthermore, Deshpande et al. \cite{8} focused on the morphological characteristics of embryos and summarized automated embryo grading methods, aiming to enhance the reproducibility and predictive efficiency of traditional subjective grading. In particular, Isa et al. \cite{blastocystreview} focused on the automated analysis of blastocyst stage embryos and summarized a series of AI-based evaluation methods, including key tasks such as blastocyst quality grading and blastocyst segmentation. In addition, Alsaad et al. \cite{14} focused on time-lapse imaging data and summarized relevant AI algorithms based on its unique temporal information.\\
\indent From a more macro perspective, Merican et al. \cite{9} and Hanassab et al. \cite{10} comprehensively sorted out the research status and development trend of gamete assessment and embryo assessment. Given that the ultimate goal of embryo grading is to optimize transfer success, some researchers are beginning to incorporate pregnancy outcome prediction into the evaluation framework. For example, Mapstone et al. \cite{11} and Louis et al. \cite{12} summarized models for embryo morphology grading and pregnancy outcome prediction. In order to further improve the comprehensive understanding of the application of ART, some efforts \cite{17}\cite{18}\cite{19}\cite{20}\cite{assessment} attempted to integrate gamete assessment (including sperm and oocytes), embryo assessment (cleavage and blastocyst stage), preimplantation genetic testing (PGT), pregnancy prediction and other tasks into the same review framework, as shown in Fig. \ref{adventage}. Among them, \cite{assessment} made a comprehensive summary of the historical development and technical path of ART technology.\\
\indent In addition to the review from the perspective of specific tasks, Dimitriadis et al. \cite{13} proposed a systematic classification method based on the embryonic development stages (including sperm stage, oocyte stage, pronuclear stage, cleavage stage and blastocyst stage), which integrated the automated tasks involved in each stage from the perspective of time. At the application level, \cite{15}\cite{16} more emphasized the potential of artificial intelligence technology in the clinical scenario of assisted reproduction, especially the possibility of combining robotics and automated operating systems to realize the whole process.\\
\indent Although prior reviews have examined the application of AI in ART, this review offers several notable advantages as shown in Table \ref{tablereview}. Compared with \cite{8}\cite{blastocystreview}\cite{9}\cite{10}\cite{14}\cite{11}\cite{12} our review encompasses a broader range of data modalities and a more comprehensive set of tasks, thereby providing a more systematic overview of the current researches. In contrast to \cite{7}\cite{17}\cite{18}\cite{13}\cite{19}\cite{20}, while we do not cover gamete assessment or gene prediction, we focus on the core tasks of embryo grading and pregnancy prediction in ART, with greater technical depth. The exclusion of gamete assessment is justified by its technical similarity to embryo grading, as many methodologies are transferable between the two. Gene prediction is omitted due to the limited number of studies and the lack of a well-established research framework in this area. Compared with \cite{assessment}\cite{15}\cite{16}, our review is confined to AI technologies and does not address robotics. This scope is deliberate, as our study is positioned at the clinical decision-making level rather than the clinical operation level, enabling a more precise response to the practical needs of clinical decision support systems.\\
\indent Furthermore, existing reviews have not yet delved deeply into the specific impact of different data modalities on task modeling and algorithm applicability. In fact, assisted reproductive practices involve various types of data, including static images, time-lapse videos, structured tabular data, etc. Thus, reorganizing tasks based on data sources can not only more accurately depict the essence of the problem but also help clarify the rationality and limitations of model design. This review is precisely based on this core motivation, focusing on the two major tasks in ART - embryo grading and pregnancy prediction. It conducts a systematic classification and comparative analysis from different data modalities. The main contributions of this paper include:
\begin{itemize}
\item This review presents a systematic review of prior research on embryo grading and pregnancy prediction, with a classification based on data modalities. Additionally, a comparative analysis is conducted on the datasets, modeling methods, and model performance reported across the reviewed studies.
\item This review systematically organizes publicly available datasets in the field of ART and explores the core challenges and future research directions in depth. Key issues include multi-modal feature fusion, data scarcity, model generalization, and the improvements of ethical frameworks.
\item This review can provide strong support for researchers to comprehensively understand the research trends of multi-modal AI in the field of ART.
\end{itemize}
\begin{figure}[t]
  % \centering
  \includegraphics[width=0.9\textwidth]{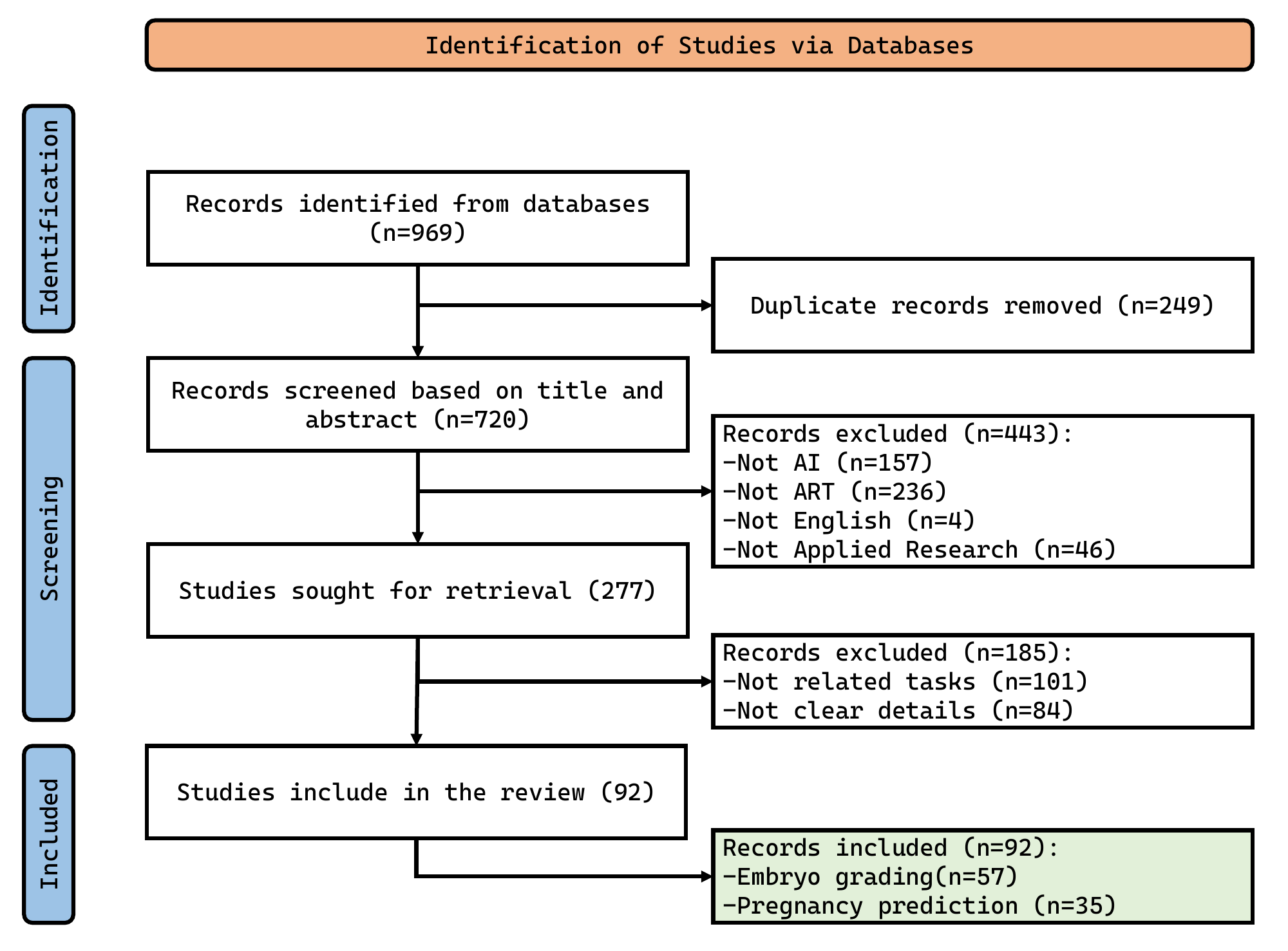}
  \caption{The process of study identification, screening, and inclusion in the review.}
  \label{database}
\end{figure}
\section{Method}\label{sec3}
This study adopted a systematic literature review approach to comprehensively integrate the existing research achievements of AI in the field of embryo grading and pregnancy prediction. The inclusion and exclusion process is shown in Fig. \ref{database}.
\subsection{Search strategy}
In terms of the selection of retrieval databases, to balance the professionalism, comprehensiveness and technical relevance of the study, the retrieval covered six core English databases. Among them, Web of Science and Scopus, as comprehensive academic databases, covered a wide range of disciplines and can cover the cross-disciplinary research of reproductive medicine and computer science, ensuring an adequate study base. PubMed, as a professional medical database, focused on the field of clinical medicine, especially enriching clinical research literature related to ART. IEEE Xplore belonged to the technical database, focusing on the application research of technologies such as AI and machine learning (ML) in medical scenarios, and facilitating the access to technology-driven study. Google Scholar and Semantic Scholar served as supplementary databases, enabling the search of preprints, conferences and other literature, while covering interdisciplinary research and reducing search blind spots.\\
\indent The search time range was set from January 1, 2020 to March 31, 2025. This move aimed to include research achievements from the past five years, fully reflecting the latest progress and research hotspots in the application of AI technology in the field, and avoiding conclusion lag due to outdated literature.\\
\indent In terms of search words, we adopted a combination of "task words + technical words". The task words included "in vitro fertilization - embryo transfer" (IVF-ET), "Assisted Reproductive Technology" (ART), "embryo grading", and pregnancy prediction. Technical words included "Artificial Intelligence" (AI), "Deep Learning" (DL), "Machine Learning" (ML), "computer vision", and "Convolutional Neural Network" (CNN). Each database independently conducted searches and only retained the top 20 items to focus on core and high-impact research within the field.\\
\indent In the supplementary search stage, to avoid missing key study, on the basis of database search, the "Reference List" and "Cited Study List" of the included literature were manually traced to supplement the original research that was in line with the theme. Through the above process, a total of 969 literature were initially retrieved.
\subsection{Inclusion and exclusion criteria}
Studies were included if they met all of the following predefined criteria: (1) the study's primary focus was the application of AI to either embryo quality grading or pregnancy outcome prediction within the context of ART, excluding gamete quality assessment or genetic testing research; (2) the publication type was an original research article, encompassing peer-reviewed journal articles and full papers from major international conferences, thereby excluding reviews, commentaries, editorials, and case reports; (3) the publication was in the English language to ensure methodological accuracy and prevent misinterpretation of technical details; (4) the study was strictly limited to human embryos, excluding all research on animal models; and (5) the study provided complete and transparent reporting of three key domains: dataset characteristics (e.g., sample size, data modality), AI model specifics (e.g., algorithm type, network architecture), and model performance metrics (e.g., Accuracy, AUC).\\
\indent The screening process was divided into three stages: The first stage was the initial screening and deduplication. Based on the title and abstract, 249 duplicate documents were removed, leaving 720. The second stage was the re-screening of topic matching. The full text of the studies was read through, and 443 studies that were "not related to AI" , "not related to ART", "not English", or "not applied research" were excluded, leaving 277. The third stage was a meticulous screening and quality verification. For the remaining 277 documents, we first excluded those with "deviated research topics" (such as gametes assessment, gene prediction, and non-human research), and then eliminated those with "incomplete information" (no details of dataset, no details of AI methods, and no performance indicators). A total of 92 studies that met the criteria were ultimately included. Classified by research topics, among the 92 studies, 57 focused on "the application of AI in embryo grading", and 35 focused on "the application of AI in pregnancy prediction".
\begin{figure}[t]
  % \centering
  \includegraphics[width=1\textwidth]{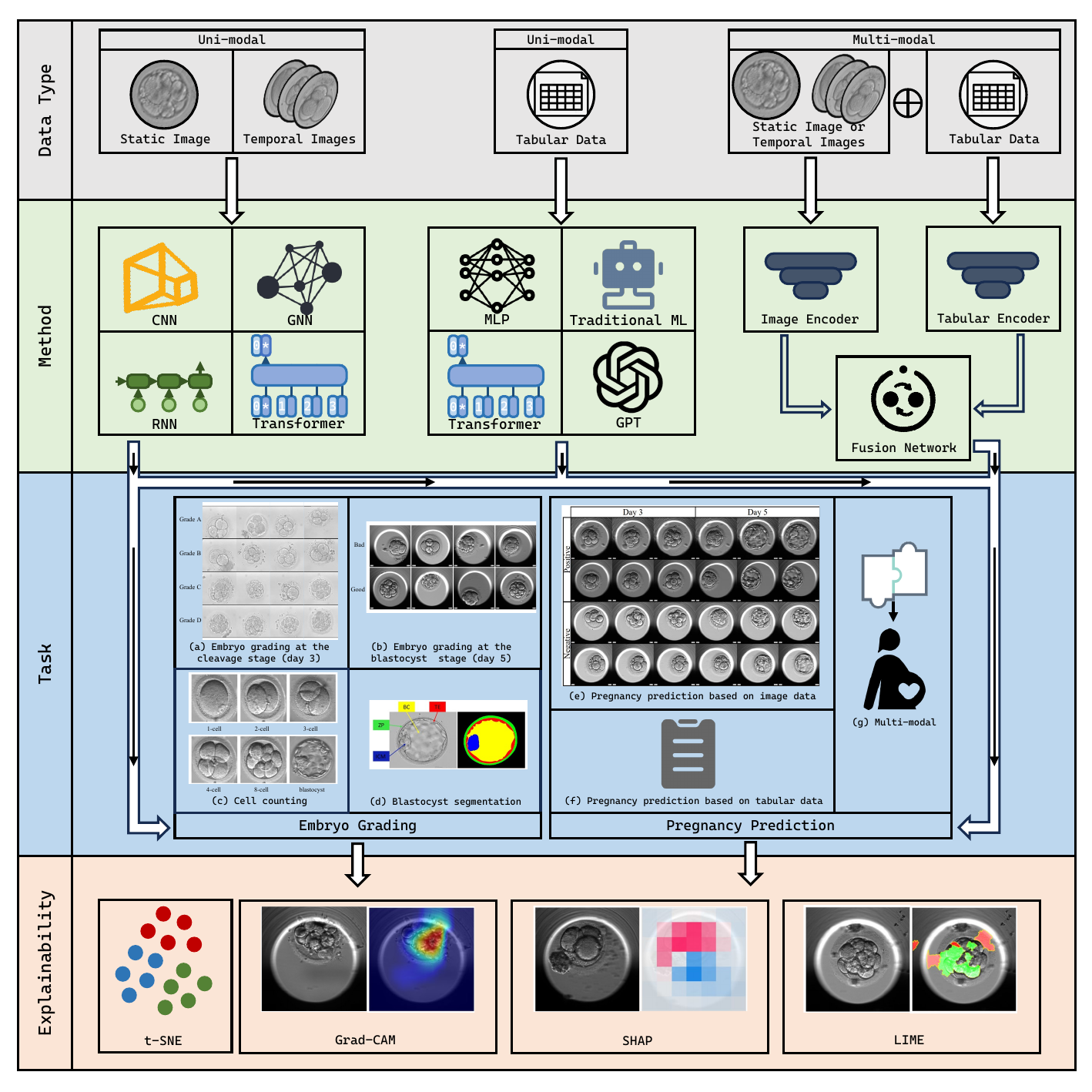}
  \caption{A flow for embryo grading and pregnancy prediction based on different data modalities and different methods. (a) Grading of cleavage-stage embryos, from high (A) to low (D) \cite{38}. (b) Grading of blastocyst-stage embryos, from high (Good) to low (Bad) \cite{49}. (c) Cell counting task, identifying the number of cells in the embryo \cite{52}. (d) Blastocyst segmentation task, segmenting different regions of the embryo \cite{65}. (e) Pregnancy prediction task based on cleavage or blastocyst embryos, positive represents successful pregnancy and negative represents failed pregnancy \cite{78}.}
  \label{method}
\end{figure}
\section{AI in ART}\label{sec4}
With the continuous development of medical imaging and machine learning technology, AI is gradually applied to the IVF-ET process, bringing new trends to reproductive medicine. In the process of IVF-ET, embryo grading and pregnancy prediction are two key steps that affect the pregnancy outcome. Traditionally, embryo grading mainly relies on manual assessment by experienced embryologists based on the morphological characteristics of embryos, while pregnancy prediction is usually based on comprehensive analysis of parental clinical characteristics (such as age, BMI, quality of sperm) and high-quality embryo image features selected by reproductive doctors.\\
\indent As a technology that can automatically learns feature representation and performs predictive analysis from large-scale and complex data, AI has shown important potential in the field of IVF-ET. Through Convolutional Neural Network (CNN), Recurrent Neural Network (RNN) and other methods, researchers can directly extract features from static embryo images, temporal embryo images, and multi-modal clinical information to effectively improve the consistency of embryo grading and the accuracy of pregnancy prediction. It should be pointed out that, as shown in the Fig. \ref{method}, the selected methods are not the same for different tasks and different data modality. Based on this, this section systematically reviews the research of AI-based embryo grading and pregnancy prediction in recent years, focusing on the current technical path, model performance, data modalities, main challenges and future development trends in clinical application, so as to provide theoretical support and practical reference for the further optimization of intelligent ART.

\subsection{Embryo grading}
Embryo grading is a critical step in ART. By assessing morphological features, it enables the selection of embryos with the highest implantation potential, thereby improving IVF success rates. Traditionally, grading has relied on manual evaluation by embryologists using criteria such as cell number, size uniformity, zona pellucida integrity, and fragmentation. With advances in computer vision, AI-based grading methods are now emerging as a promising research direction.\\
\indent Depending on the type of input data, the embryo grading tasks can be divided into two main categories: grading based on static embryo images and grading based on temporal embryo images. The former is usually evaluated using a single image of an embryo developing to a specific stage, such as day 5. While the latter analyzes image sequences of the embryo at multiple consecutive time points, such as day 1 to day 5, to capture its dynamic development process. As shown in Fig. \ref{method}, in the "Task" column: Fig. \ref{method}(a) corresponds to the embryo grading task based on the static cleavage stage image (day 3). Embryos are divided into four grades: A (high), B, C, and D (low). The higher the grade, the more regular the arrangement of cells in the embryo. Fig. \ref{method}(b) corresponds to the embryo grading task based on the static blastocyst stage image (day 5). Embryos are classified into two grades: "Good" (high) and "Bad" (low), and the higher the grade, the more complete the development of the blastocyst. In addition, the dataset for embryo grading task based on temporal embryo images is shown in Fig. \ref{step}(d). The input of the dataset is a series of images, and the labels are classified according to the embryo development level, just like static embryo images.\\
\indent It is important to note that embryo grading tasks are often closely linked to other related tasks, such as cell counting and embryo segmentation. Embryo development grade is closely related to the number of cells. Therefore, accurate cell counting is essential for embryo grading. At the same time, the embryo segmentation task aims to divide the embryo into different regions, such as the ICM and TE, whose morphological characteristics are of great reference value for assessing the embryo. By analyzing these derived tasks in depth, a more complete understanding of the developmental state of the embryo can be achieved, thereby improving the accuracy and reliability of embryo grading. As shown in Fig. \ref{method}, in the "Task" column: Fig.\ref{method} (c) corresponds to the cell counting task; Figure.\ref{method} (d) corresponds to the blastocyst segmentation task.

\begin{sidewaystable}[htbp]
  \centering
  
  \caption{Datasets, methods and performance under embryo grading task and cell counting task (image-based).}
    \begin{tabular}{l|l|l|l|l|l}
    \hline
    \hline
    \textbf{Paper} & \textbf{Dataset (number, type)} & \textbf{Year} & \textbf{Method} & \textbf{Task} & \textbf{ACC} \\
    \hline
    Septiandri\cite{24} & 1084, day3 image (static) & 2020  & ResNet50 & 3-embryo grading & 91.79\% \\
    Mohamed\cite{26} & 110, day5 image (static) & 2023  & VGG16 & 4-embryo grading & 90.00\% \\
    Wu\cite{28} & 3601, day3 image (static) & 2021  & CNN+Ensemble & 4-embryo grading & 74.14\% \\
    Wu\cite{31} & 3599, day3 image (static) & 2022  & CNN+Attention & 4-embryo grading & 71.06\% \\
    Perumal\cite{33} & 840, day3 and day5 image (static) & 2024  & GAT   & 2-embryo grading & 98.80\% \\
    Liu\cite{35} & 273438, day2-day6 image (static) & 2025  & Transformer & 16-embryo grading & 73.10\% \\
    Wicaksono
\cite{grading11} & 1226, day3 image (static) & 2021  & ResNet50+SSL & 3-embryo grading & 80.30\% \\
    Paya
\cite{43} & 3014, day4, day5 image (static) & 2022  & ResNet50+SSL & 2-embryo grading & 87.17\% \\
    You\cite{36} & 5736, day3 image (static) & 2025  & UNet+GCN & 4-embryo grading & 86.96\% \\
    Guo\cite{37} & 7728, day3 image (static) & 2022  & Grad-CAM+ResNet & 4-embryo grading & 78.61\% \\
    Guo\cite{38} & 7728, day3 image (static) & 2023  & UNet+CNN & 4-embryo grading & 79.56\% \\
    Sariniya\cite{grading6} & 220, day3 image (static) & 2025  & Morphology+EfficientNet & 4-embryo grading & 94.30\% \\
    Wang\cite{41} & 1025, day5 image (11 focal) & 2021  & VGG16 & 2-embryo grading & 89.90\% \\
    Liu\cite{44} & 4997, day5 image (7 focal) & 2024  & ResNet50+Attention & 10-embryo grading & 86.53\% \\

    Nguyen\cite{grading3} & 1135, day3 image (static) & 2021  & VGG16 & 2-embryo grading & 75.00\% \\
    Wu\cite{grading8} & 8660, day5 image (static) & 2024  & ResNet50 & 6-embryo grading & 94.00\% \\
    Vermilyea
\cite{grading9} & 8886, day3 image (static) & 2020  & ResNet152 & 5-embryo grading & 64.30\% \\
    Alkindy\cite{grading1} & 152, day3 image (static) & 2023  & ResNet+Xception & 4-embryo grading & 98.00\% \\
    Misaghi\cite{grading4} & 273438, day2-day6 image (static) & 2024  & EfficientNet & 17-embryo grading & 87.00\% \\
    Thirumalaraju\cite{42} & 2440, day5 image (static) & 2020  & Ensemble & 5-embryo grading & 64.96\% \\
    Iza\cite{grading2} & 249, day5 image (static) & 2024  & ResNet50+Attention & 9-embryo grading & 69.80\% \\
    Kim\cite{grading5} & 14989, day3 image (static) & 2024  & ResNet50+Multiscale & 2-embryo grading & 95.89\% \\
    Sariniya\cite{grading7} & 200, day3 image (static) & 2025  & Morphology+VGG16 & 2-embryo grading & 88.10\% \\
    Wang\cite{grading12} & 41279, day1-day6 image (static) & 2024  & Transformer+SSL & 2-embryo grading & 82.30\% \\

    Abbasi\cite{45} & 127, day5 image (temporal) & 2021  & ResNet18+Concat & 2-embryo grading & 73.08\% \\
    Abbasi\cite{46} & 130, day3 and day5 image (temporal) & 2021  & CNN+Add & 2-embryo grading & 76.90\% \\
    Kragh\cite{47} & 4032, day5 image (temporal) & 2019  & InceptionV3+LSTM & 6-embryo grading & 67.40\% \\
    Vaidya\cite{21} & 803, day 2,3 and 5 image (temporal) & 2021  & VGG16+LSTM & 5-embryo grading & 100.00\% \\
    Kalyani\cite{48} & 704, day1-day3 image (temporal) & 2024  & ResNet50+GRU & 2-embryo grading & 93.05\% \\
    Wang\cite{49} & 531, day1-day3 image (temporal) & 2023  & R(2+1)D & 2-embryo grading & 77.74\% \\
    Barhoun
\cite{temgrading1} & 704, day1-day5 image (temporal) & 2025  & R(2+1)D & 2-embryo grading & 93.30\% \\
    Fordham\cite{51} & 136, day1-day5 image (temporal) & 2022  & Transformer & 3-embryo grading & 62.50\% \\
    Canat\cite{temgrading2} & 1909, day1-day5 image (temporal) & 2024  & Transformer & 11-embryo grading & 66.30\% \\
    Shi\cite{temgrading3} & 23027, day1-day5 image (temporal) & 2025  & Transformer(spatial+Temporal) & 9-embryo grading & 63.50\% \\
    Sun\cite{temgrading4} & 2586, day1-day5 image (temporal) & 2025  & Transformer(spatial+Temporal) & 3-embryo grading & 86.06\% \\
    
    Malmsten\cite{52} & 661062, day1-day5 image (static) & 2020  & InceptionV3 & 8-cell counting & 93.90\% \\
    Nguyen\cite{53} & 148918, day1-day3 image (static) & 2022  & Transformer & 4-cell counting & 94.30\% \\
    Zhou\cite{54} & 442, day3 image (static) & 2022  & ResNet50+Genetic algorithm & 20-embryo grading & 86.41\% \\

    Zhou\cite{counting1} & 1500, day2-day5 image (static) & 2025  & FPN+Attention & 8-cell counting & 86.92\%(f1score) \\
    Dong\cite{coungting2} & 4805, day3 image (static) & 2024  & YOLOv7 & 4-cell counting & 92.40\%(mAP50) \\
    \hline
    \end{tabular}%
  \label{table1}%
\end{sidewaystable}

\subsubsection{Static embryo grading}
When grading embryos based on static embryo images, the current processing strategy was generally using CNN to extract image features and then grade embryos. Septiandri et al. \cite{24} used ResNet50 \cite{25} to perform a 3-grading task on the 3rd day of embryo development, and the final classification Accuracy (ACC) could reach 91.79$\%$. Similarly, Mohamed et al. \cite{26} used VGG16 \cite{27} to perform a 4-grading task on the 5th day of embryonic development, and the final ACC was 90.00$\%$. In order to combine the advantages of different CNNs, Wu et al. \cite{28} collected 3601 images on the 3rd day of embryo development of 1800 couples for 4-grading, and extracted the features of DenseNet169 \cite{29}, Inception V3 \cite{30}, ResNet50 and VGG19 last layer for ensemble classification, respectively. And the final ACC was 74.14$\%$. Attention mechanism, as a module to enhance feature extraction, was often used in combination with CNN to improve the model's attention to key regions during embryonic development. Wu et al. \cite{31} added channel-level attention and point-level attention to CNN, enabling the model to adaptively focus on regions related to the embryo grade. Eventually, the ACC of embryo 4-grading on the test set was 71.06$\%$, and the ablation experiments also demonstrated the effectiveness of the attention mechanism.\\
\indent In addition to CNN, some recent studies have also begun to introduce Graph Neural Network (GNN) and Transformer \cite{32} into embryo grading tasks. For example, Perumal et al. \cite{33} directly converted images into graph nodes and input them into GAT \cite{34}, and extracted features for embryo 2-grading through the message passing mechanism of GNN. Liu et al. \cite{35} converted images into patches, obtained local features through self-attention operation within patches, and obtained global features through Transformer among patches, which were finally used for embryo 16-grading.\\
\indent In addition to architectural improvements, \cite{grading11}\cite{grading12} explored training strategy innovations by introducing contrastive learning into the embryo quality grading task. Their approach leverages a self-supervised learning (SSL) framework \cite{contra1}\cite{contra2}, where discriminative and generalizable feature representations were learned through instance discrimination by constructing positive and negative sample pairs. By eliminating the reliance on manually annotated labels, this method effectively enhanced the performance of downstream grading tasks, demonstrating the potential of self-supervised representation learning in embryo image analysis.\\
\indent While single-stage models offered simplicity and efficiency, they often struggled to balance accuracy and computational cost in complex tasks. To address this, two-stage models had been widely adopted as a more refined alternative. The first stage performed coarse screening to generate candidate regions, while the second stage refined these candidates through fine-grained classification. You et al. \cite{36} and Guo et al. \cite{37}\cite{38} extracted the cell segmentation map and Grad-CAM \cite{39} attention map of embryos in the first stage, respectively, so that the model in the second stage could focus on regions of interest with embryo grading. Sariniya et al. \cite{grading6} employed two branches, one obtaining features through CNN and the other through morphological analysis, and ultimately fused them for embryo grading.\\
\indent In addition, since the microscope could take multi-angle embryo images with different focal lengths, some studies had begun to explore methods for embryo grading after fusion of these multi-angle images \cite{41}\cite{44}. This multi-angle image fusion strategy was expected to further improve the accuracy and reliability of embryo grading.\\
\indent The core characteristics and implementation details of the aforementioned studies are summarized in TABLE \ref{table1}. To broaden the scope and enhance the comprehensiveness of this review, we further incorporate a number of related works \cite{grading3}\cite{grading8}\cite{grading9}\cite{grading1}\cite{grading4}\cite{42}\cite{grading2}\cite{grading5}\cite{grading7}\cite{grading12} that share similar technical methodologies or target comparable application tasks. This enables a more systematic and comparative analysis, facilitating a holistic understanding of the current research landscape.

\subsubsection{Temporal embryo grading}
In addition to basic microscopy, Time-Lapse Microscopy (TLM) was often used to observe embryos. This microscope was capable of taking TLM images, which were a set of temporally sequenced embryo images as shown in Fig. \ref{step}(d). TLM could capture the changing process of cells over time, showing the process of cell division, morphological change, movement, development, etc. Compared with the static image, TLM images had an additional time dimension. In order to solve the problem of processing multiple time-series images, Abbasi et al. \cite{45}\cite{46} extracted the features of a single image by CNN, and directly fused the features of multiple images by addition or concatenation operation for the subsequent embryo grading task. Apart from these simple and direct methods, Kragh et al. \cite{47}, Vaidya et al. \cite{21} and Kalyani et al. \cite{48} extracted single image features through InceptionV3, VGG16 or ResNet50, and then integrated the features of TLM images through RNN for embryo internal quality grading.\\
\indent To effectively process three-dimensional time series data, Wang et al. \cite{49} and Barhoun\cite{temgrading1} adopted R(2+1)D network \cite{50} (a 3DCNN method) to extract features from dynamic embryo images and applied it to the 2-embryo grading task. Due to the natural advantages of Transformer model in processing temporal data, Fordham et al. \cite{51} and Canat\cite{temgrading2} directly converted temporal images into tokens and input them into Transformer for blastocyst classification. To better exploit the spatial structural information within embryo images and the temporal dynamics across image sequences, Shi et al. \cite{temgrading3} and Sun et al. \cite{temgrading4} proposed Transformer-based architectures for spatial and temporal feature extraction. By leveraging the self-attention mechanism to model long-range dependencies, these approaches effectively capture the spatiotemporal evolution patterns inherent in embryonic development, enabling a more comprehensive understanding of dynamic morphological changes.\\
\indent The core characteristics and implementation details of the aforementioned studies are summarized in TABLE \ref{table1}.

\subsubsection{Cell counting}
Although embryo grading could directly reflect the embryo development potential, it needed professional embryologists to manually label, and the cost was high. Cell counting had become a research hotspot because of its relatively convenient label acquisition and the number of cells can indirectly characterize the quality of embryos. For example, Malmsten et al. \cite{52} achieved high-precision counting of 8-cell embryos based on the improved InceptionV3 model, with an accuracy of 93.90$\%$. Nguyen et al. \cite{53} further introduced the time dimension and fused continuous developmental image sequences through the Transformer model, which significantly improved the counting accuracy. Aiming at the interference of exfoliated cells in embryo grading, Zhou et al. \cite{54} proposed to extract regions of interest in the preprocessing stage to eliminate interference, and combined with adaptive genetic algorithm to optimize network parameters, successfully realized the integrated processing of cell counting and embryo grading.\\
\indent In addition to image classification-based approaches, object detection techniques were also extensively employed for embryonic cell counting. For example, Zhou et al. \cite{counting1} integrated an attention mechanism with a Feature Pyramid Network \cite{fpn} to perform detection and enumeration of embryonic cells. Similarly, Dong et al. \cite{coungting2} utilized the YOLOv7 model \cite{yolov7}, achieving a mAP@0.5 score of 92.4$\%$, which demonstrates excellent performance in this task. The core characteristics and implementation details of the aforementioned studies are summarized in TABLE \ref{table1}.

\begin{table*}[htbp]
  \centering
  \caption{Datasets, methods and performance under embryo segmentation task (image-based).}
  \resizebox{0.95\textwidth}{!}{
    \begin{tabular}{l|l|l|l|l|l}
    \hline
    \hline
    {\multirow{2}{*}{\textbf{Paper}}} & {\textbf{Dataset}} & {\multirow{2}{*}{\textbf{Year}}} & {\multirow{2}{*}{\textbf{Method}}} & {\textbf{Region of}} & {\multirow{2}{*}{\textbf{Dice}}} \\
     & \textbf{(number, type)} &  &  & \textbf{segmentation} & \\
    \hline
    \multirow{2}{*}{Saeedi\cite{55}} & \multirow{42}{*}{211, day5 image (static)} & \multirow{2}{*}{2017} & K-means+ & TE    & 77.30\% \\
   &  &       & Watershed algorithm & ICM   & 83.10\% \\
   \multirow{2}{*}{Harun\cite{56}} &  & \multirow{2}{*}{2019} & \multirow{2}{*}{UNet} & TE    & 92.50\% \\
    &   &       &   & ICM   & 94.30\% \\
   Rad\cite{57} &   & 2020  & Dilated U-Net & TE    & 86.61\% \\
  \multirow{2}{*}{Corpuz\cite{58}} &   & \multirow{2}{*}{2023} & \multirow{2}{*}{ResUNet} & TE    & 91.20\% \\
     &   &       &   & ICM   & 98.70\% \\
   \multirow{2}{*}{Chen\cite{59}} &   & \multirow{2}{*}{2025} & \multirow{2}{*}{Embryo-Net} & TE    & 81.20\% \\
    &   &       &   & ICM   & 86.51\% \\
    \multirow{2}{*}{Vullota\cite{seg3}} &  & \multirow{2}{*}{2025} & \multirow{2}{*}{DeepLab} & TE    & 85.00\% \\
    &   &       &   & ICM   & 87.00\% \\
   \multirow{3}{*}{Miled\cite{60}} &   & \multirow{3}{*}{2025} & \multirow{3}{*}{ResA-Net} & TE    & 81.80\% \\
     &   &       &   & ICM   & 83.09\% \\
     &   &       &   & ZP    & 86.87\% \\
  \multirow{5}{*}{Rad\cite{61}} &   & \multirow{5}{*}{2019} & \multirow{5}{*}{Blast-Net} & TE    & 76.52\% \\
     &   &       &   & ICM   & 81.07\% \\
    &   &       &   & BC    & 80.79\% \\
    &   &       &   & ZP    & 81.15\% \\
     &   &       &   & BG    & 94.74\% \\
   \multirow{5}{*}{Mushtaq\cite{62}} &   & \multirow{5}{*}{2022} & \multirow{5}{*}{ECS-Net} & TE    & 78.43\% \\
    &   &       &   & ICM   & 85.26\% \\
    &   &       &   & BC    & 88.41\% \\
   &   &       &   & ZP    & 85.34\% \\
    &   &       &   & BG    & 94.87\% \\
  \multirow{5}{*}{Arsalan\cite{63}} &   & \multirow{5}{*}{2022} & \multirow{5}{*}{SSS-Net} & TE    & 78.15\% \\
    &   &       &   & ICM   & 84.50\% \\
    &   &       &   & BC    & 88.68\% \\
    &   &       &   & ZP    & 84.51\% \\
   &   &       &   & BG    & 95.82\% \\
    \multirow{5}{*}{Arsalan\cite{64}} &   & \multirow{5}{*}{2022} & \multirow{5}{*}{MASS-Net} & TE    & 79.08\% \\
    &   &       &   & ICM   & 85.88\% \\
     &   &       &   & BC    & 89.28\% \\
   &   &       &   & ZP    & 84.69\% \\
    &   &       &   & BG    & 96.07\% \\
  \multirow{5}{*}{Ishaq\cite{65}} &   & \multirow{5}{*}{2023} & \multirow{5}{*}{FSBS-Net} & TE    & 80.17\% \\
    &   &       &   & ICM   & 85.55\% \\
     &   &       &   & BC    & 89.15\% \\
     &   &       &   & ZP    & 85.80\% \\
      &   &       &   & BG    & 95.62\% \\
   \multirow{5}{*}{Arsalan\cite{66}} &   & \multirow{5}{*}{2024} & \multirow{5}{*}{PSF-Net} & TE    & 80.00\% \\
    &   &       &   & ICM   & 86.46\% \\
    &   &       &   & BC    & 90.15\% \\
     &   &       &   & ZP    & 85.77\% \\
    &   &       &   & BG    & 96.10\% \\
    \hline
    Jamal\cite{67} & 327, day3 image (static) & 2023  & DenseUNet & Cell  & 96.62\% \\
    \hline
    Zhao\cite{seg2} & 1218, day1 image (static) & 2021  & UNet & Cell  & 86.90\% \\
    \hline
    Zhang\cite{68} & 1548, day3 image (static) & 2024  & SAM   & Fragments & 69.50\% \\
    \hline
    Liao\cite{seg1} & 551, day2, day3 image (static) & 2024  & Faster RCNN+Unsupervised   & Cell & 88.00\% \\
    \hline
    \end{tabular}%
  \label{table2}%
  }
\end{table*}%

\subsubsection{Embryo segmentation}
In embryo assessment, the task of embryo segmentation was equally important. Image segmentation of embryos at the cleavage stage and blastocyst stage aimed to accurately identify and quantify different structures and cell compositions during embryonic development, thereby providing a reliable basis for embryo quality assessment in ART. At the blastocyst stage, segmentation mainly focused on Inner Cell Mass (ICM), TrophEctoderm (TE), BlastCoele (BC) and Zona Pellucida (ZP) to support more detailed blastocyst grading and PGT biopsy. Saeedi et al. \cite{55} was the early work that proposed automated embryo segmentation, using K-means and watershed algorithm to segment ICM and TE, and the final Dice scores were 83.10$\%$ and 77.30$\%$, respectively. In addition, this work published relevant dataset, a total of 211 embryo images with segmentation labels, which promoted the development of this field. Based on the previous work, many studies \cite{56}\cite{57}\cite{58} began to segment ICM and TE by improving UNet \cite{80}. In addition, Chen et al. \cite{59} proposed a segmentation model based on spatial modeling based on the prior knowledge that the ICM is in the center and the TE is in the periphery. In addition to the segmentation of ICM and TE, there are many studies \cite{60}\cite{61}\cite{62}\cite{63}\cite{64}\cite{65}\cite{66} which had included BC, ZP and background (BG) regions into the segmentation criteria.\\
\indent Although the segmentation of blastocyst stage embryos was the mainstream, there were also some works segmenting cleavage stage embryos. At the cleavage stage, segmentation helped to extract key morphological features such as cell number, symmetry, and cytoplasmic distribution, which are closely related to the developmental potential of the embryo. Jamal et al. \cite{67} used DenseUNet to segment the cells by collecting and labeling images of 327 embryos on the third day of development, and the final Dice score was 96.62$\%$. Zhao et al. \cite{seg2} segmented the ZP, cytoplasm and pronucleus on the first day of embryonic development, and the final comprehensive dice score was 86.90$\%$. Zhang et al. \cite{68} introduced a SAM-based \cite{69} two-branch segmentation method, in which the semantic branch was used to segment fragments, and the instance branch was used to detect and segment cleavage stage embryos. The final mAP on the embryo cells was 87.40$\%$, and the Dice score on the fragments was 69.50$\%$. Liao et al. \cite{seg1} proposed a serial processing strategy, which first used the Faster R-CNN model to detect the target of embryonic cells, and then further used the unsupervised segmentation method to achieve accurate cell segmentation based on the positioning information provided by the detection results.\\
\indent The core characteristics and implementation details of the aforementioned studies are summarized in TABLE \ref{table2}.

\begin{sidewaystable}[htbp]
  \centering
  \caption{Datasets, methods and performance under pregnancy prediction (image-based, table-based and multi-modal-based).}
    \begin{tabular}{l|l|l|l|l}
    \hline
    \hline
    \textbf{Paper} & \textbf{Dataset (number, type)} & {\textbf{Year}} & \textbf{Method} & \textbf{Score} \\
    \hline
    Xu\cite{70} & 185, day3 image (static) & 2014  & LBP+SVM & ACC=64.71\% \\

    Miyagi\cite{preg1} & 160, day5 image (static) & 2019  & LR & AUC=65.00\% \\

    Geller\cite{73} & 361, day5 image (static) & 2021  & InceptionV1 & AUC=65.70\% \\

    Sawada\cite{74} & 470, day5 image (static) & 2021  & ResNet50+Attention & AUC=64.20\% \\

    Rad\cite{preg2} & 578, day5 image (static) & 2019  & DeepLab+ResNet50 & AUC=70.90\% \\

    Berntsen\cite{75} & 14644, day1-day5 image (temporal) & 2021  & I3D+LSTM & AUC=67.00\% \\

    Abbasi\cite{78} & 130,day3 and day5 image (temporal) & 2023  & ResNet50+Transformer & AUC=77.50\% \\

    Borna\cite{79} & 252, day 1,2,3 image (temporal) & 2024  & UNet+InceptionV3 & AUC=75.00\% \\

    Ouyang\cite{95} & 4046, day1,2,3 image (temporal) & 2025  & Transformer & AUC=61.70\% \\

    Mapstone\cite{81} & 700, day1-5 image (temporal) & 2024  & MobileNet & AUC=68.00\% \\

    Boucret\cite{preg3} & 1580, day1-5 image (temporal) & 2025  & SSL+LSTM+XGBoost & AUC=64.00\% \\

    Nagaya\cite{preg4} & 643, day1-5 image (temporal) & 2022  & ResNet50+Attention & AUC=65.40\% \\

    Liu\cite{82} & 401, 19 tabular data & 2021  & RF    & AUC=61.30\% \\

    Mehta\cite{86} & 333, 11 tabular data & 2024  & AdaBoost & AUC=100.0\% \\

    Dehghan\cite{88} & 812, 25 tabular data & 2024  & AdaBoost+Genetic algorithm & AUC=91.00\% \\

    Mohammad\cite{89} & 10036, 46 tabular data & 2025  & LR & AUC=92.12\% \\

    Sergeev\cite{90} & 8732, 19 tabular data & 2024  & MLP   & AUC=79.80\% \\

    Gong\cite{91} & 657, 25 tabular data & 2023  & LR    & AUC=87.90\% \\

    Zhu\cite{92} & 969, 23 tabular data & 2024  & LR    & AUC=76.50\% \\

    Wu\cite{93} & 15779, 11 tabular data & 2024  & RF    & AUC=71.60\% \\

    Cao\cite{94} & 2865, 123 tabular data & 2024  & GPT   & AUC=87.00\% \\

    Ouyang\cite{95} & 4046, 22 tabular data & 2025  & Transformer & AUC=71.30\% \\

    Borji\cite{tab1} & 665244, 94 tabular data & 2024  & Transformer & AUC=99.96\% \\

    Liu\cite{tabcnn} & 48514, 38 tabular data & 2025  & CNN & AUC=88.99\% \\

    Cordeiro\cite{tab2} & 183, 180 tabular data & 2025  & RF & AUC=99.65\% \\

    Gao\cite{tab3} & 13826, 16 tabular data & 2021  & LR & AUC=74.30\% \\

    Goyal\cite{tab4} & 495630, 94 tabular data & 2020  & RF & AUC=86.40\% \\

    Liu\cite{tab5} & 1857, 48 tabular data & 2023  & LGBM & AUC=63.40\% \\

    Wen\cite{tab6} & 949, 20 tabular data & 2022  & XGBoost & AUC=78.70\% \\

    Enatsu\cite{96} & 19342, day5 image (static)+13 tabular data & 2022  & ResNet18+RF & AUC=71.00\% \\

    Charnpinyo\cite{97} & 1099, day5 image (static)+4 tabular data & 2023  & EfficientNet+MLP & AUC=72.00\% \\

    Salih\cite{multi2} & 1394, day5 image (static)+17 tabular data & 2025  & ResNet34+MLP & AUC=91.00\% \\

    Kim\cite{99} & 2555, day5 image (static)+1 tabular data & 2024  &  Image segmentation+ResNet50+MLP & AUC=74.10\% \\

    Liu\cite{100} & 17580, day5 image (static)+16 tabular data & 2023  & EfficientNet+MLP & AUC=77.00\% \\

    Kim\cite{101} & 3286, day1-day5 image (temporal)+ unknown tabular data & 2024  & Transformer & AUC=68.80\% \\

    Ouyang\cite{95} & 4046, day1,2,3 image (temporal)+22 tabular data & 2025  & Transformer+DeFusion & AUC=74.60\% \\

    Duval\cite{multi3} & 9982, day1-day5 image (temporal)+32 tabular data & 2023  & 3DResNet+XGBoost & AUC=72.70\% \\
    \hline
    \end{tabular}%
  \label{table3}%
\end{sidewaystable}%

\subsection{Pregnancy prediction}
Although deep learning has made great progress in the task of embryo grading and has surpassed that of humans, the purpose of embryo grading is for pregnancy prediction. The role of embryo grading in pregnancy prediction is indirect, so current research tends to directly predict pregnancy outcomes. Pregnancy prediction is to predict the success of pregnancy after embryo transfer by analyzing the parents' clinical data, embryo characteristics data, and laboratory test data. Current AI-based pregnancy prediction tasks can be divided into image-based, table-based and multi-modal based tasks according to different data types. Image-based task can be further divided into static image-based and temporal image-based. One point to be stated is that in this section we classify all three types of pregnancy, hCG biochemical pregnancy, fetal heart clinical pregnancy, and live birth, as pregnancies. As shown in Fig. \ref{method}, in the "Task" column, Fig.\ref{method} (e) represents the negative and positive embryos for pregnancy prediction based on the third day or the fifth day.

\subsubsection{Pregnancy prediction based on static image}
The earlier machine learning method for pregnancy prediction based on static embryo images could be traced back to 2014. Xu et al. \cite{70} extracted the features of static embryo images to 256 dimensions by Local Binary Pattern (LBP) \cite{71}. Then Support Vector Machine (SVM) \cite{72} was used to classify the extracted features, and the comprehensive performance of the test data in six institutions was ACC=64.71$\%$. Miyagi et al. \cite{preg1} collected 180 blastocyst images and used a Logistic Regression (LR) model combined with five-fold cross-validation to predict pregnancy outcomes, achieving an Area Under receiver operating characteristic Curve (AUC) of 65.00$\%$. Geller et al. \cite{73} collected 361 blastocyst stage embryo images and used the InceptionV1 network to predict pregnancy. The final AUC was 65.70$\%$. Sawada et al. \cite{74} fused the static embryo image information extracted by CNN branch and attention mechanism branch for pregnancy prediction, and the AUC reached 64.20$\%$. Rad et al. \cite{preg2} employed a two-stream network to extract the features of the blastocyst segmentation image and the original image respectively, and then fused them for pregnancy prediction. The final accuracy rate was 70.90$\%$.

\subsubsection{Pregnancy prediction based on temporal image}
Previous image-based pregnancy prediction studies generally used images of the last day of embryo development to predict pregnancy, which ignored the timing of embryo development. Therefore, pregnancy prediction studies based on temporal embryo images had been gradually developed. Berntsen et al. \cite{75} used I3D network \cite{76} (a 3DCNN method) to obtain the features of 128 sequences of embryo development images, and then reduced the feature dimension by pooling operation. Finally, bidirectional LSTM \cite{77} was used to connect temporal features for the final pregnancy prediction.\\
\indent Unlike image analysis, video analysis requires spatial-temporal data analysis, but many methods for processing spatial-temporal data such as 3DCNN are not suitable for medical applications due to high computational complexity. Therefore, Abbasi et al. \cite{78} divided the pregnancy prediction task of temporal embryo image into three stages. In the first stage, CNN was used to extract individual image features on day 3 and day 5, and in the second stage, the temporal image features were fused through the attention mechanism in Transformer. In the third stage, the results of day 3 and day 5 time-series images were integrated by voting method to make the final pregnancy prediction. When processing TLM images, many frames of images would be recorded for each embryo, but these images had a lot of redundant information because the changes of embryos were small in a short time. Borna et al. \cite{79} only used the three key development images of the first, second and third days of embryo development for pregnancy prediction. The ACC of the model reached 75.00$\%$, far higher than 60.32$\%$ of professional embryologists. Also using keyframe temporal images, Ouyang et al. \cite{95} proposed a spatial-temporal position encoding, making the Transformer more suitable for fusing temporal images. In order to understand the process of embryo development more comprehensively, Mapstone et al. \cite{81} carried out the embryo grading task and pregnancy prediction task at the same time, and finally the ACC of the embryo 5-grading task was 88.00$\%$, and the AUC of pregnancy prediction was 68.00$\%$.\\
\indent Boucret et al. \cite{preg3} first extracted the video features based on the self-supervised learning method, and then used XGBoost. The final AUC was 64.00$\%$. In clinical practice, the label of pregnancy negative was less dependent on the quality of the embryo, which would cause the result of model overfitting when training. To solve this problem, Nagaya et al. \cite{preg4} used the method of unlabeled learning to change the label of pregnancy negative to unlabeled, which enhanced the regularization of the model and improved the performance of the model.

\subsubsection{Pregnancy prediction based on tabular data}
In clinical practice, reproductive doctors needed to predict pregnancy outcomes not only based on embryo images, but also based on tabular data such as parental fertility indicators. Tabular data is usually structured and easily quantified, and did not require the collection of large amounts of data like image processing tasks. Liu et al. \cite{82} collected 401 cases of data with 19 fertility indicators, such as female age, female BMI, and EndoMetrial Thickness (EMT), and used SVM, LR \cite{83}, Decision Tree (DT) \cite{84}, and Random Forest (RF) \cite{85} to predict pregnancy. The AUC of the best RF method was 61.30$\%$, and it was found through statistics that: Among patients with good-quality embryos, the clinical pregnancy success rate was about 70$\%$ when EMT thickness was greater than 9.6 mm, but 50$\%$ when EMT thickness was no greater than 9.6 mm. Mehta et al. \cite{86} collected data of 333 cases with 11 indicators such as female age, Antral Follicle Count (AFC), and Follicle-Stimulating Hormone (FSH), and observed that the ACC of pregnancy prediction of Adaboost \cite{87} was 97.50$\%$, better than all other machine learning models, and found statistically that: The propensity to clinical pregnancy was negative if the woman was older than 36 years, and the likelihood of clinical pregnancy decreased substantially if the woman was older than 40 years; Furthermore, the propensity for clinical pregnancy was positively correlated with the number of embryos transferred in the same IVF cycle.\\ 
\indent Similarly, through statistical related studies \cite{88}\cite{89}\cite{90}\cite{91}\cite{92}\cite{93}, it can be concluded that the characteristics affecting pregnancy prediction mainly include female age, Anti-Mullerian Hormone (AMH), EMT, semen volume, oocytes number, AFC, female BMI and so on. In addition to using traditional machine learning methods for pregnancy prediction, Cao et al. \cite{94} converted 123 relevant data into natural language and fine-tuned the large language model GPT-4 with the converted natural language, resulting in a final AUC of 87.00$\%$. Ouyang et al. \cite{95} and Borji \cite{tab1} et al. converted the tabular data into tokens and input them into Transformer for pregnancy prediction. Liu et al. \cite{tabcnn} transformed one-dimensional tabular data into two-dimensional data through replication operations and then extracted features through CNN for pregnancy prediction.\\
\indent The core characteristics and implementation details of the aforementioned studies are summarized in TABLE \ref{table3}. To broaden the scope and enhance the comprehensiveness of this review, we further incorporate a number of related works that share similar technical methodologies or target comparable application tasks \cite{tab1}\cite{tab2}\cite{tab3}\cite{tab4}\cite{tab5}\cite{tab6}.

\subsubsection{Pregnancy prediction based on multi-modal data}
With the progress of AI and the improvement of medical data collection technology, pregnancy prediction was more likely to use multi-modal data such as images and tables. This trend was mainly due to the potential of multi-modal data fusion, which integrated the advantages of different types of data to provide more comprehensive prediction results, thereby improving the clinical application value of pregnancy prediction. Compared with image data, tabular data could directly express high-level semantics information. Therefore, some works \cite{96}\cite{97}\cite{multi2} extract image information into low-dimensional data through CNN, and then input it into the fusion model combined with tabular data directly. The quality of the embryo at the blastocyst stage is mainly reflected from two aspects, ICM and TE. Based on this prior knowledge, Kim et al. \cite{99} first binarized and segmented the ICM and TE of the embryo through the traditional image segmentation method, and then added the segmentation map to the original image to obtain the embryo enhancement image. Finally, the features extracted from the enhancement image were combined with age features in the pregnancy prediction task. Similarly, Liu et al. \cite{100} collected 17580 multi-modal data for pregnancy prediction, in which the image data contained two images focusing on ICM and TE respectively, and the tabular data contained 16 parental fertility indicators.\\
\indent Considering that the image modality data used in the previous multi-modal pregnancy study were all static, Kim et al. \cite{101} collected the time series embryos image data from day 1 to day 5 and health spreadsheet data. When processing the image data, it was multi-stage. Firstly, each morphology of the embryo was segmented through the segmentation model. Then the segmentation images were input into the neural network to predict five key indicators: blastocyst size, blastocyst level, cell boundary, cell number and developmental stage. After processing, the original time series images, spreadsheet data, morphological segmentation images and five key indicators were input into Transformer for final pregnancy prediction. In addition, Ouyang et al. \cite{95} were the first to integrate images of the first 3 days of embryo development and fertility table data for pregnancy prediction, and proposed an effective decoupling fusion network to fuse multi-modal data. Dual et al. used 3DResNet to extract video data and combined it with 31 tabular data for the final pregnancy assessment. The details of these methods are shown in the TABLE \ref{table3}.
\begin{figure}[t]
  % \centering
  \includegraphics[width=1.0\textwidth]{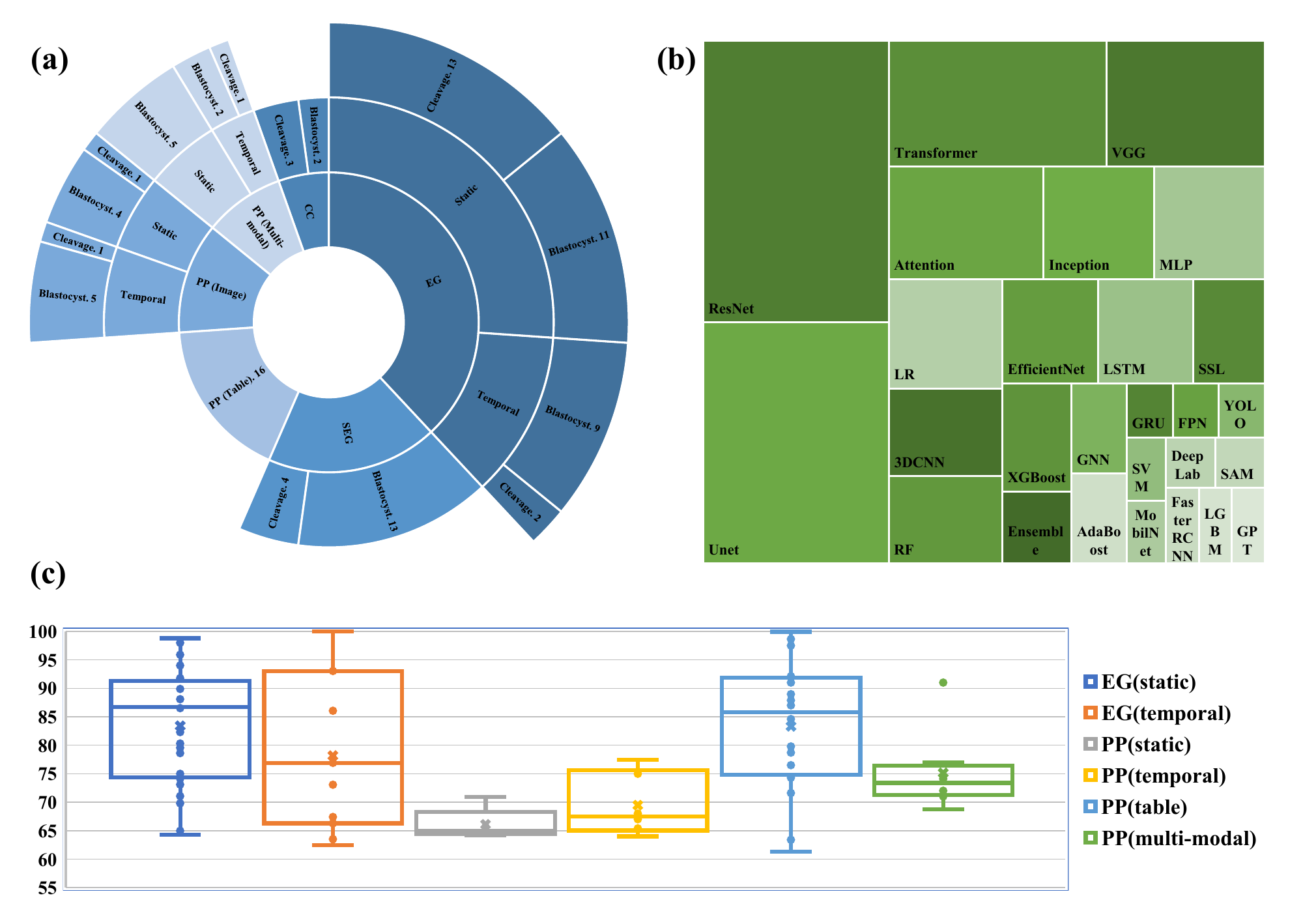}
  \caption{(a) A sunburst chart of different periods (cleavage stage and blastocyst stage), different modalities (static images, temporal images, tables and multi-modalities), and different task types (embryo grading (EG), cell counting (CC), embryo segmentation (ES) and pregnancy prediction (PP)). (b) A tree diagram of AI technology in the ART field: High-frequency technologies stand out, the size of the technology area reflects the usage frequency, and some traditional or unique methods are not included in the statistics. (c) A box plot of model performance across different modalities and tasks (for the EG task, performance is evaluated by ACC, whereas for the PP task, the AUC is the metric ).}
  \label{perfomance}
\end{figure}

\subsection{Summary of current works}
Fig. \ref{perfomance}(a) visually presents the distribution relationships among different periods (cleavage stage and blastocyst stage), different modalities (static images, temporal images, tables and multi-modalities), and different task types (embryo grading (EG), cell counting (CC), embryo segmentation (ES) and pregnancy prediction (PP)) in the form of a sunburst chart. It can be clearly seen from the figure that the number of studies related to embryo grading is significantly greater than that of pregnancy prediction tasks, and the research work at the blastocyst stage is far more than that at the cleavage stage. This distribution feature indicates that the research related to pregnancy prediction during the cleavage stage is relatively weak. Future studies may consider strengthening the exploration in these directions.\\
\indent Fig. \ref{perfomance}(b) delineates the technical methodologies AI in ART through a tree diagram. Synthesizing this visualization with the current studies reveals several prominent technical routes, which are primarily determined by the data modality.  For static image analysis, the predominant approach involves feature extraction using foundational deep learning architectures, including CNNs, Transformers, and GNNs.  In contrast, the analysis of temporal image sequences necessitates models capable of capturing dynamic information, such as RNNs, 3D CNNs, and Transformer-based architectures. For structured tabular data, conventional yet highly efficient machine learning algorithms are typically employed. In addition, some methods take a different approach, using language models or CNNs to handle tabular modal data.\\
\indent Beyond these foundational models, several advanced strategies are consistently utilized to augment predictive performance. Ensemble learning and attention mechanisms have proven effective in significantly boosting model accuracy. Furthermore, to address the inherent complexity of ART tasks, researchers often decompose problems into manageable sub-tasks using strategies like incorporating prior medical knowledge, implementing two-stage models, or designing multi-network frameworks. Finally, emerging techniques such as self-supervised learning show considerable promise for enhancing model representational ability, particularly in research contexts constrained by limited data availability.\\
\indent Fig. \ref{perfomance}(c) presents the statistical distribution of performance for different tasks in the form of a box plot. Due to the differences in the evaluation criteria of embryo grading tasks, the performance variance is relatively large. However, the average ACC of embryo grading under both static images and temporal images exceeds 75$\%$, and it has significantly outperformed the subjective assessment level of embryologists \cite{36}\cite{37}\cite{grading3}.\\
\indent At present, a variety of mature products have been put into clinical use, further promoting the standardization and intelligence of embryo grading in the field of assisted reproduction. STORK (https://github.com/ih-lab/STORK) is an AI model from Stanford University, trained on the Inception image recognition architecture. The model can identify the morphological characteristics of the embryo and output the transfer probability or classification results. Other commercial software scoring systems, such as KIDScore and iDAScore \cite{108}, are also widely used. They are fully AI-driven scoring systems developed by Vitrolife, outputting a continuous score (e.g., 1 to 10) for prioritization of embryos. Meanwhile, ERICA (Embryo Ranking Intelligent Classification Algorithm) \cite{109} is also an embryo evaluation model based on deep learning, aiming to automatically score and rank static embryos in ART, which can be accessed on https://embryoranking.com/.\\
\indent In addition, when conducting pregnancy prediction tasks based on tabular data, due to the significant differences in the clinical features included by different hospitals or institutions, coupled with the non-uniformity of testing equipment, data collection methods, and standardized processing procedures, the model performance shows a large variance as shown in Fig. \ref{perfomance}(c). At present, there is no recognized standardized process for processing tabular data, which to some extent limits the generalization ability and clinical applicability of the model among different centers. And the pregnancy prediction model based on the table's uni-modal model demonstrates superior predictive performance compared to the multi-modal model. This phenomenon may be related to the differences in the number of indicators used by different institutions: models based on tabular data use an average of 48 indicators, while multi-modal models use an average of only 15 indicators. The significant difference in the number of indicators may be one of the key factors leading to performance differences.\\
\indent Although the tabular indicators adopted in different studies vary, in the task of pregnancy prediction, by comparing and analyzing the statistical performance of static images, temporal images and multi-modal methods as shown in Fig. \ref{perfomance}(c), it can be clearly observed that the multi-modal method introducing tabular modality has significantly better performance than the method relying only on a single image modality (static or temporal). This result indicates that the clinical features contained in the tabular data are complementary to the image data. Fusing multi-modal data can effectively enhance the prediction accuracy and robustness of the model. The above findings are highly consistent with the results of \cite{95}\cite{96}\cite{100}. To advance the clinical implementation of pregnancy prediction, it is imperative to establish standardized processing methodologies for tabular data modalities, integrate multi-modal data, and subsequently develop high-performance, highly reliable models leveraging large-scale datasets.

\section{Explainability}\label{sec5}
\indent In ART, AI models are widely applied in embryo quality assessment, pregnancy success prediction and other key diagnosis and treatment tasks. Although these models show significant advantages in performance, their "black box" feature makes explainability analysis be a prerequisite for clinical deployment. On the one hand, ART involves highly sensitive issues such as bioethics, patients' right to informed consent, and the attribution of medical responsibility. The decision-making basis of the model must be transparent to doctors and patients. On the other hand, explanatory analysis not only helps enhance doctors' trust in the prediction results of AI, but also helps identify potential medical-related features.\\
\indent From the perspective of specific methods, t-distributed Stochastic Neighbor Embedding (t-SNE) \cite{40} can be used for the visualization of high-dimensional feature spaces. By reducing the dimension of the embryo image features extracted by the deep model and mapping them to the two-dimensional plane, the clustering structure of embryo samples in the representation space can be revealed, which indirectly verifies whether the model has learned discriminative semantic features \cite{42}\cite{43}.\\
\indent To address the interpretability challenge inherent in image-based models, researchers have developed a range of visualization analysis techniques, as illustrated in Fig. \ref{method}. Among these, Gradient-weighted Class Activation Mapping (Grad-CAM) has emerged as a widely adopted technique. It operates by generating a heatmap that visually highlights the spatial regions upon which the model focuses its attention. This capability allows researchers to verify whether the AI system is attending to clinically significant structures within the embryo image, such as the inner cell mass (ICM) or the trophectoderm (TE), thereby enabling a critical evaluation of the rationale and clinical validity of its decision-making process \cite{41}\cite{112}. Beyond Grad-CAM, other prominent methods offering similar functionalities include Local Interpretable Model-agnostic Explanations (LIME) \cite{lime} and SHapley Additive exPlanations (SHAP) \cite{98} . These approaches further bolster the transparency and trustworthiness of model decisions by employing local approximation or feature attribution analysis, respectively \cite{explainall}.\\
\indent Besides image models, SHAP and LIME are also suitable for interpretability analysis of tabular data. These two methods can quantitatively evaluate the contribution of each input variable to the model prediction results, and clarify the positive or negative influence of key factors such as patient age, hormone level, and embryo score on the output probability and their weight distribution \cite{95}\cite{96}. Therefore, SHAP and LIME show significant advantages in processing multi-modal data, which can effectively explain the feature interactions from different modalities, and provide transparent and reliable explanation support for the decision-making process of complex models.

\begin{table}[htbp]
  \centering
  \caption{Descriptions of different public datasets, including the source of the dataset, the modality of the dataset, the applicable tasks, and the download links. }
    \begin{tabular}{m{4em}|m{2em}|m{4em}|m{25em}}
    \hline
    \hline
    \textbf{Dataset} & \textbf{Year} & \textbf{Modality} & \textbf{Description} \\
    \hline
    \cite{102}  & 2022  & Static image & The dataset consists of 704 time-lapse videos of developing embryos with 7 focal length, covering day 2-6 of embryo development, for a total of 2.4 million images. Annotations are graded to 16 different developmental stages. \url{https://zenodo.org/records/6390798} \\
    \hline
    \cite{51}  & 2022  & Temporal image & The dataset consists of 143 time-lapse videos of embryo development, including day 1-5 of embryo development. Annotations are 36 score ratings annotated by embryologists, which can be further divided into three grades: good, medium, and bad. \url{https://dataverse.harvard.edu/dataset.xhtml?persistentId=doi:10.7910/DVN/1Z4HLC} \\
    \hline
    \cite{103} & 2022  & Static image & The dataset consists of 2440 images, including day 5 images of embryo development. The annotation is the five ratings 1-5, where ratings 1 and 2 are for no blastocysts developed and ratings 3,4 and 5 are for blastocysts developed. \url{https://osf.io/3kc2d/} \\
    \hline
    \cite{104} & 2023  & Static image & The dataset consists of 1020 images, including images on day 3 and day 5 of embryo development. The annotations are good and bad ratings. \url{https://kaggle.com/competitions/world-championship-2023-embryo-classification} \\
    \hline
    \cite{105} & 2025  & Static image & The dataset consists of 5500 images generated by AI, including images from day 1-5 of embryo development. The annotations are the five grades 2-cell, 4-cell, 8-cell, morula, and blastocyst. \url{https://zenodo.org/records/14253170}, \url{https://huggingface.co/datasets/deepsynthbody/synembryo\_latentdiffusion}, \url{https://huggingface.co/datasets/deepsynthbody/synembryo\_stylegan} \\
    \hline
    \cite{55} & 2017  & Static image & The dataset consists of 211 images, including day 5 images of embryo development. Annotations are segmented labels for different regions of the blastocyst, namely ICM, TE, BC, and ZP. \url{https://vault.sfu.ca/index.php/s/066vGJfviJMYuP6/authenticate} \\
    \hline
    \cite{68} & 2024  & Static image & The dataset consists of 1548 images, including images from day 1 to 3 of embryo development. Annotation is the COCO object detection format, which detects objects separately for embryos, fragments, and background. \url{https://www.kaggle.com/datasets/austin012/cleavageembryo-dataset/data} \\
    \hline
    \cite{106} & 2024  & Table & The dataset consists of 8732 treatment cycles and includes 19 key tabular data. Label is the result of fetal heart clinical pregnancy. \url{https://github.com/embryossa/KAN-in-IVF} \\
    \hline
    \cite{107} & 2023  & Static image and Table & The dataset consists of 2344 images from 837 cases, including images on day 4 or 5 of embryo development. These 2344 images were annotated with Expansion 5-grading, ICM 3-grading, and TE 3-grading. In addition, the 837 cases included six tabular features for hCG biochemical pregnancy, fetal heart clinical pregnancy, and live birth annotation. \url{https://figshare.com/articles/figure/Blastocyst\_dataset\_zip/20123153/3?file=39348899} \\
    \hline
    \end{tabular}%
  \label{table4}%
\end{table}%

\section{Public Dataset}\label{sec6}
Since medical image data usually involve patient privacy information and legal compliance requirements, public datasets may violate relevant regulations or cannot pass ethical review, so most ART-related datasets have not been opened to the public. Considering the importance of open data to promote the development of medical artificial intelligence and computer-assisted therapy technology, this paper summarizes the currently publicly available representative datasets as listed in TABLE \ref{table4}.\\
\indent Although some datasets in the field of ART have been published, their distribution and quality are significantly limited. Specifically, most of the existing public datasets focus on the embryo grading task, which has been maturely applied and widely implemented in clinical practice, and its research value tends to be saturated \cite{108}. In contrast, the dataset of embryo segmentation task is not only released earlier, but also limited in sample size, which is difficult to meet the requirements of current deep learning models for data quantity and quality. In addition, for the critical clinical task of multi-modal pregnancy prediction, only a small dataset is available \cite{107}, which is far from enough to support the training and validation of the model. Therefore, there is an urgent need to construct high-quality and large-scale pregnancy prediction datasets to promote the scientific research and clinical translation application of related algorithms.
\section{Challenge and Future Work}\label{sec7}
\begin{figure}[t]
  % \centering
  \includegraphics[width=0.9\textwidth]{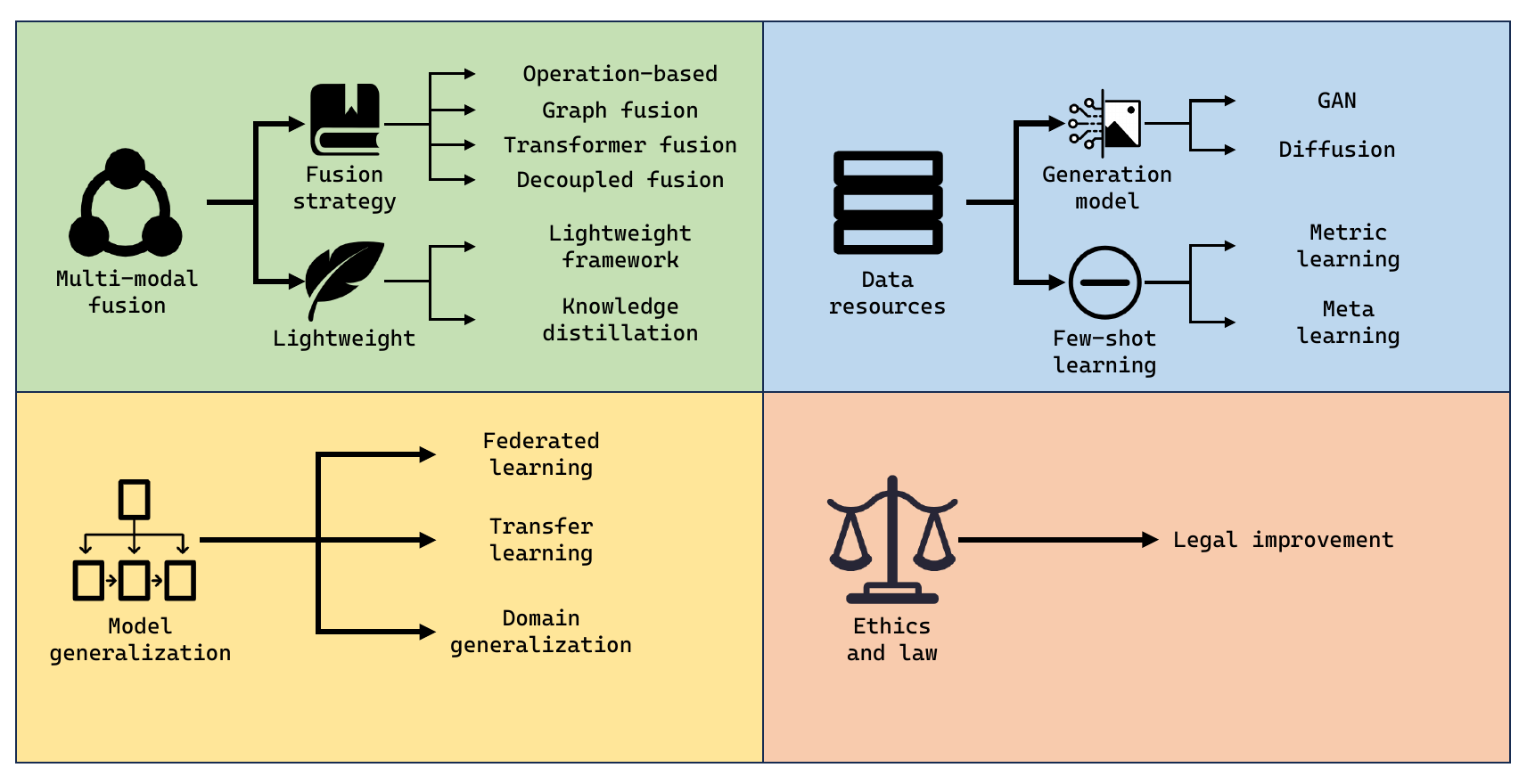}  \caption{Current challenge and future work.}
  \label{challenge}
\end{figure}

The application of AI in ART has made remarkable progress in recent years, covering many key tasks from gamete quality assessment, embryo grading to pregnancy outcome prediction. Although previous studies have fully demonstrated the great potential of AI in improving assessment consistency, prediction accuracy and process automation, its clinical deployment still faces multiple challenges. In the following, the challenges of AI in the field of ART are systematically analyzed from four dimensions: multi-modal fusion, data resources, model generalization, and ethical barriers as shown in Fig. \ref{challenge}.

\subsection{Multi-modal fusion}
As discussed in Section 4.3, the task of embryo grading has reached a state of maturity, achieving successful clinical translation \cite{108}. The sign of embryo grading maturity lies in the fact that its performance has surpassed that of professional embryologists \cite{36}\cite{37}\cite{grading3}. Consequently, the focus of contemporary research is progressively shifting towards the more formidable challenge of predicting pregnancy outcomes. The inherent biological complexity of pregnancy, governed by a multitude of interacting factors, cannot be fully captured by models reliant on uni-modal data, such as static embryo images. This limitation constitutes a fundamental performance bottleneck. Therefore, the integration of multi-modal information—encompassing static images, time-lapse videos, and structured clinical and laboratory data—has emerged as a pivotal strategy and a frontier direction for enhancing the predictive accuracy of these models.\\
\indent Despite its considerable potential, the practical implementation of multi-modal fusion strategies is impeded by two core bottlenecks. The first challenge stems from the heterogeneity and semantic gap inherent in multi-modal data \cite{clip}. Static images, time-lapse videos, and structured tabular data originate from distinct feature spaces and exhibit significant disparities in their intrinsic semantic representations and information densities. A naive approach, such as simple feature concatenation or early fusion, often results in a “forced alignment” that fails to achieve effective cross-modal synergy. Instead, this method risks introducing noise and degrading model performance due to semantic conflicts or information redundancy between modalities, a phenomenon known in the literature as negative transfer \cite{negativetrasfer}.\\
\indent The second bottleneck pertains to the architectural complexity and substantial computational overhead of multi-modal models \cite{computation}. These models typically necessitate parallel feature extraction networks and sophisticated fusion modules, resulting in an architecture that is significantly larger and more complex than their uni-modal counterparts. This complexity translates into a formidable demand for computational resources, including high-performance GPUs, large memory capacity, and protracted training times.    This high computational cost not only constrains the efficiency of model development and iterative optimization but also poses a significant barrier to the practical deployment of such models in clinically resource-constrained environments.\\
\indent To address the aforementioned challenges, future research in pregnancy prediction for ART should prioritize two key directions: the innovation of multi-modal fusion strategies and the architectural design of lightweight models.
\subsubsection{Multi-modal Fusion Strategies}
Traditional fusion methods, such as simple feature concatenation or element-wise operations, are computationally efficient but fail to capture the complex, non-linear interactions inherent in heterogeneous ART data \cite{116}\cite{113}. To overcome this limitation, more sophisticated paradigms are required. GNNs offer a powerful framework by modeling multi-modal entities as nodes in a graph, enabling the explicit mining of high-order semantic relationships through message-passing mechanisms \cite{114}. Similarly, the Transformer architecture, with its self-attention and cross-attention mechanisms, excels at capturing long-range dependencies within individual modalities (e.g., time-lapse videos) and facilitating precise cross-modal information alignment \cite{115}. Furthermore, decoupled fusion presents a promising approach to mitigate negative transfer by decomposing multi-modal features into modality-invariant and modality-specific subspaces, thereby preserving complementary information while minimizing interference \cite{95}\cite{117}\cite{118}\cite{119}.
\subsubsection{Model Lightweight}
While these advanced fusion strategies enhance predictive performance, they often introduce substantial computational complexity, hindering deployment in resource-constrained clinical settings. Therefore, model lightweight is imperative. A primary approach involves employing lightweight backbone networks \cite{120}\cite{121}, such as MobileNet for image data or simplified MLPs for tabular data, which utilize efficient architectural designs like depth-wise separable convolutions to reduce the parameter count and computational load. A more sophisticated technique is knowledge distillation, where a compact “student” model is trained to get the ability of a larger, high-performing “teacher” model \cite{122}\cite{mmkd}. The resulting highly accurate and efficient models are essential for enabling real-time AI applications in clinical ART.
\subsection{Data resources}
As discussed in Section 6, data resources in the field of ART exhibit significant scarcity characteristics due to their inherent sensitivity to human privacy, ethical review protocols, and complex sample collection processes. This data dilemma precipitates two interrelated challenges that critically impede the robust iteration and objective evaluation of AI models in ART domain.\\
\indent Firstly, the lack of reproducibility and benchmarking undermines scientific progress. At present, the vast majority of published research results are based on the private data sets of each institution. Due to the differences in data distribution, collection standards and processing procedures, it is difficult to compare and evaluate the experimental results between different studies, which greatly weakens the universality of research conclusions. Especially, the field of ART still lacks a large-scale, high-quality, standardized public multi-modal dataset, and this "data gap" has become a key bottleneck for AI to move from algorithm research to clinical application in this field.\\
\indent Secondly, the adaptation of existing models on private small datasets is insufficient. When faced with limited scale and expensive labeling of clinical data, directly transferring models pre-trained on general large-scale datasets, or relying on simple data augmentation techniques, often yields little benefit. These methods are difficult to fundamentally solve the problem of overfitting caused by insufficient data, so it is urgent to develop more sophisticated algorithm strategies designed for small sample scenarios.\\
\indent To address the above challenges, academia and industry have explored a number of parallel and complementary technical paths. At the data level, although it is the most ideal solution to manually collect and construct large-scale public datasets through multi-center cooperation, it is often difficult to be widely promoted in the short term due to legal barriers to data sharing, interest coordination between institutions, and long ethical approval processes. Thus, leveraging generative models to "create" data becomes a highly potential alternative. For example, generative adversarial networks \cite{123} and emerging diffusion models \cite{124} can learn the intrinsic distribution of real data and generate highly realistic and diverse synthetic embryo images, thus effectively expanding the size and diversity of the training set and alleviating the problem of data scarcity under the premise of protecting patient privacy. There have been relevant works \cite{105}\cite{generatedata} applying generation technology to the field of ART.\\
\indent At the algorithmic level, few-shot learning provides a fundamental paradigm for solving the small sample size problem. It enables the model to "learn how to learn" by introducing mechanisms such as metric learning or meta-learning \cite{125}. The combination of these advanced algorithm strategies provides practical technical support for breaking through the data barriers in the field of ART and promoting the clinical implementation of AI models.

\subsection{Model generalization}
\indent Due to the differences in data acquisition equipment, scoring standards, and patient population characteristics between different laboratories, AI models often face the problem of performance degradation in the cross-institution deployment process, which exposes the limitations of insufficient generalization ability. \\
\indent In order to improve the cross-domain adaptation ability of the model, the future should focus on the development of transfer learning \cite{126} and domain adaptation \cite{127} technology to realize the dynamic adjustment of model parameters or feature space. At the same time, as a collaborative training framework with local data, federated learning \cite{128} provides a feasible path for breaking data islands and realizing cross-institution model sharing. For example, Wang et al. \cite{fl} proposed a federated learning framework combined with an expectation maximization algorithm for image assessment of embryonic cleavage quality, which enhanced prediction accuracy while protecting privacy.

\subsection{Ethics and law}

AI's decision-making participation in ART, especially when it involves sensitive tasks such as embryo scoring, genetic screening, and pregnancy success prediction, raises a series of ethical and legal issues. For example, the scoring of embryos may be seen as judging the "potential value of life". In addition, AI intervention in the field of reproduction may also have a subtle impact on society's conception of fertility and family structure ethics. Among them, the most prominent challenge is the question of responsibility \cite{law9}. With the continuous digitalization of the healthcare system and the wide application of AI algorithms in clinical decision support, more and more non-traditional medical agents, including technology developers, data providers, and algorithm operators, are gradually involved in the medical service chain, thus reshaping the subject structure of medical responsibility \cite{law4}. However, when an AI system leads to misdiagnosis, treatment delay, or other adverse clinical consequences due to algorithmic flaws or data bias, the existing legal framework faces a significant dilemma in defining who is responsible: should the developer, medical institution, clinical user, or multiple parties share the responsibility? Consensus has not yet emerged on this issue \cite{law5}.\\
\indent To complicate matters, most advanced AI models (especially deep learning systems) have a "black box" nature that lacks transparency and traceability in their decision-making processes \cite{law6}. Although Explainable Artificial Intelligence (XAI) methods have made certain progress in recent years, their explanations often stay at the surface correlation, and it is difficult to reveal the causal mechanism inside the model. This inherent opacity not only weakens clinicians' ability to understand and judge the output results of AI, but also shakes their trust in the system, which affects the effectiveness and safety of human-computer collaborative decision-making \cite{law7}.\\
\indent In addition, the behavioral deviation in human-computer interaction further aggravates the ethical and legal risks. Research shows that experienced clinicians tend to be cautious or even skeptical of AI recommendations and tend to retain independent judgment; Junior providers, on the other hand, may rely too much on system outputs and suffer from "automation bias," which is to blindly trust technology over clinical intuition and comprehensive assessment \cite{law8}. This cognitive difference not only affects the quality of diagnosis and treatment, but also introduces uncertainty in the interaction between human and algorithm in the responsibility chain, which makes the accident imputation more complex.\\
\indent Therefore, it is urgent to establish a systematic ethical evaluation framework and technical supervision mechanism to clarify the responsibility of AI in assisted reproduction. And relevant practitioners should develop norms covering data privacy protection, informed consent, model interpretability, and transparency of use. Although nascent initiatives for ethical guidelines and policy-making have begun to emerge, the governance frameworks for artificial intelligence technologies significantly lag behind their rapid advancement in the medical domain \cite{law1}. Concurrently, there is a prevailing lack of comprehensive awareness within the medical community regarding the complex ethical challenges posed by these emerging AI technologies \cite{law2}. Consequently, the risk assessment of AI-powered medical tools should not be confined solely to algorithmic accuracy. Instead, it necessitates a more holistic and context-specific evaluation. For instance, the U.S. Food and Drug Administration (FDA) typically mandates that approved AI medical devices be maintained in a “locked” state, wherein their algorithms are prohibited from being modified post-market \cite{law3}. This regulatory paradigm serves to reinforce product safety and stability.
\section{Conclusion}\label{sec8}
This review provides an in-depth discussion of the potential and challenges of artificial intelligence in embryo grading and pregnancy prediction in ART. By analyzing AI models from different data sources, we find that this technique not only effectively reduces the error caused by human factors, but also reveals the key factors affecting embryonic development through the processing of large amounts of data. However, there are still obstacles to the application of AI in ART, such as the difficulty of multi-modal data fusion and the scarcity of high-quality data. Future research should aim to address these issues and further optimize AI algorithms to adapt to different clinical needs. This will help to improve the success rate of IVF-ET, reduce the psychological stress of patients, and bring the hope of new life to more families.

\backmatter

% \bmhead{Supplementary information}

% If your article has accompanying supplementary file/s please state so here. 

% Authors reporting data from electrophoretic gels and blots should supply the full unprocessed scans for key as part of their Supplementary information. This may be requested by the editorial team/s if it is missing.

% Please refer to Journal-level guidance for any specific requirements.

\bmhead{Acknowledgements}
This work is supported in part by the Guangdong Provincial Natural Science Foundation (2023A1515011431), the Guangzhou Science and Technology Planning Project (202201010092), the National Natural Science Foundation of China (72074105).
\bmhead{Conflict of Interest}
The authors declare that they have no conflict of interest.
\bmhead{Author Contributions}
Xueqiang Ouyang: Contributed to the literature survey, conceptualization, original draft preparation, and visualization. Jia Wei: Contributed to the supervision, manuscript review and editing, and final approval of the version to be published.
\bmhead{Clinical trial number}
Not applicable.

\section*{List of Abbreviations}
\resizebox{0.7\linewidth}{!}{
\begin{tabular}{ll}
ACC   & Accuracy \\
AFC   & Antral Follicle Count \\
AI   & Artificial Intelligence \\
AMH   & Anti-Mullerian Hormone \\
ART   & Assisted Reproductive Technology \\
AUC   & Area Under receiver operating characteristic Curve \\
BC   & BlastCoele \\
BG   & Background \\
BMI   & Body Mass Index \\
CC   & Cell Counting \\
CNN   & Convolutional Neural Network \\
EG   & Embryo Grading \\
EMT   & EndoMetrial Thickness \\
GNN   & Graph Neural Network \\
Grad-CAM   & Gradient-weighted Class Activation Mapping \\
hCG   & Human Chorionic Gonadotropin \\
ICM   & Inner Cell Mass \\
ICSI   & IntraCytoplasmic Sperm Injection \\
IVF-ET   & In Vitro Fertilization-Embryo Transfer \\
LIME & Local Interpretable Modelagnostic Explanations\\
PGT   & Preimplantation Genetic Testing \\
PP   & Pregnancy Prediction \\
RNN   & Recurrent Neural Network \\
SHAP   & SHapley Additive exPlanations \\
SSL   & Self-Supervise Learning \\
TE   & TrophEctoderm \\
TLM   & Time-Lapse Microscopy \\
t-SNE   & t-distributed Stochastic Neighbor Embedding \\
ZP   & Zona Pellucida \\
\end{tabular}}

% \section*{Declarations}

% Some journals require declarations to be submitted in a standardised format. Please check the Instructions for Authors of the journal to which you are submitting to see if you need to complete this section. If yes, your manuscript must contain the following sections under the heading `Declarations':

% \begin{itemize}
% \item Conflict of interest/Competing interests (check journal-specific guidelines for which heading to use)
% \item Ethics approval and consent to participate
% \item Consent for publication
% \item Materials availability
% \end{itemize}

% \noindent
% If any of the sections are not relevant to your manuscript, please include the heading and write `Not applicable' for that section. 

% \bigskip

% Springer journals and proceedings: \url{https://www.springer.com/gp/editorial-policies}

% Nature Portfolio journals: \url{https://www.nature.com/nature-research/editorial-policies}

% \textit{Scientific Reports}: \url{https://www.nature.com/srep/journal-policies/editorial-policies}

% BMC journals: \url{https://www.biomedcentral.com/getpublished/editorial-policies}
% \end{flushleft}
\begin{appendices}

\section{Background}\label{secA1}
With the development of society and the change of lifestyle, infertility has attracted more and more attention. The emergence and development of ART has provided hope for many couples who are unable to conceive naturally. The birth of the world's first test-tube baby in 1978 marked the beginning of ART \cite{111}. In recent years, with the continuous progress of ART technology, remarkable achievements have been made in improving the pregnancy rate, reducing complications and personalized treatment, which largely provides more reproductive choices for patients with infertility. In the development process of ART technology, it can be divided into the first generation of IVF-ET, the second generation of IntraCytoplasmic Sperm Injection (ICSI), and the third generation of Preimplantation Genetic Testing (PGT). This section provides a brief introduction to the technology at each stage of ART.

\subsection{The first generation——IVF-ET}
\begin{figure}[t]
  \centering
  \includegraphics[width=1.0\textwidth]{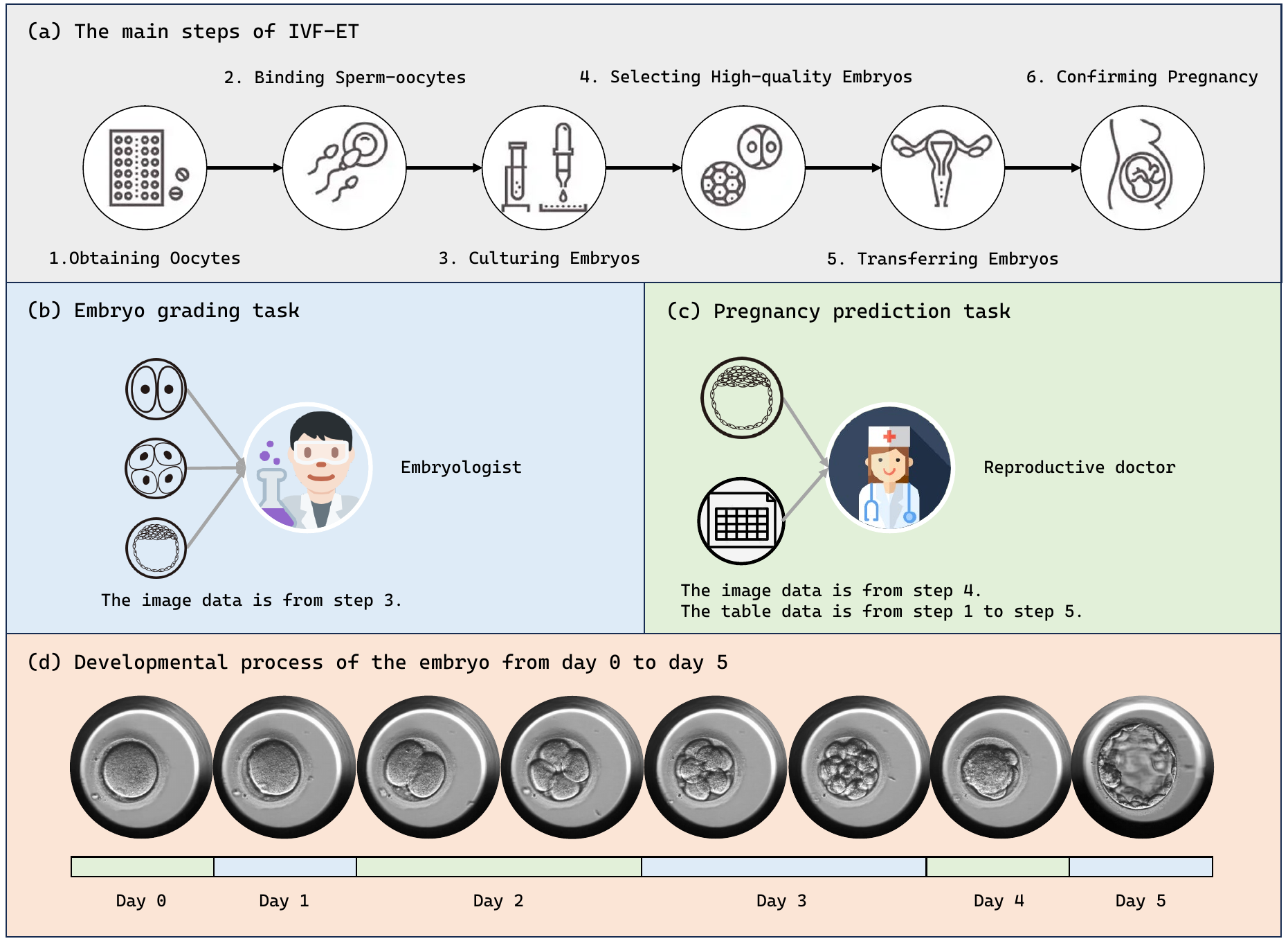}
  \caption{ (a) The main steps of IVF-ET. (b) and (c) Data modalities and completed tasks used by embryologists and reproductive doctors, respectively. (d) The development process of the embryo from day 0 to day 5.}
  \label{step}
\end{figure}

IVF-ET is currently the most widely used method in ART, which mainly targets infertility caused by female factors such as tubal obstruction and ovulation disorder. Louise Brown, the world's first test-tube baby born through IVF-ET, was born in Britain in 1978, marking a major breakthrough in the field of human assisted reproduction \cite{110}. The process of IVF-ET is shown in the Fig. \ref{step}(a).
\begin{itemize}
\item Step 1: By controlling ovarian ovulation with gonadotropin injection, we can obtain multiple mature oocytes.
\item Step 2: By collecting sperms and purifying them, we then allow the sperms to combine with oocytes in vitro to obtain zygotes.
\item Step 3: The zygotes (embryos) are continuously cultured and dynamically monitored in a constant temperature incubator at 37°C, usually for 3 to 5 days.
\item Step 4: Based on the morphological characteristics, we can select high-quality embryos with high developmental potential.
\item Step 5: The selected embryo is transferred into the mother's uterus.
\item Step 6: Pregnancy is confirmed by three methods. Biochemical pregnancy, about 10 days after embryo transfer, confirms whether the embryo implants by detecting the level of Human chorionic gonadotropin (hCG), which is an early sign of pregnancy. Fetal heart clinical pregnancy, 5 to 6 weeks after embryo transfer, an ultrasound examination is conducted to confirm whether the embryo in the uterus has a fetal heart beating, which is a mid-pregnancy sign. Live birth, whether the newborn shows vital signs after delivery, is the ultimate sign of pregnancy.
\end{itemize}
\indent In the overall IVF-ET process, high-resolution microscopic imaging can be used to capture continuous embryo development image data in the step 3, and clinical tabular data related to parental fertility can be simultaneously collected from step 1 to step 5. These tabular data include, but are not limited to: Female age, Body Mass Index (BMI), Endometrial Thickness (EMT), Antral Follicle Count (AFC), quality of sperm, etc. \\
\indent In the step 3, it is very important to continuously observe the image of embryo development for judging embryo quality and improving the success rate \cite{21}. The process of the first 5 days of embryonic development is shown in the Fig. \ref{step}(d).
\begin{itemize}
\item Day 0: The sperm enters the oocyte and the genetic material of the sperm and oocyte combine to form the zygote.
\item Day 1: The zygote starts the first cleavage, forming a 2-cell embryo.
\item Day 2: The embryo continues cleavage at the 2-4 cell stage.
\item Day 3: The embryo continue cleavage at the 6-8 cell stage.
\item Day 4: The embryo continues to divide, forming a 16-32 cell morula, where cells begin to join tightly and individual cell boundaries become blurred to form a dense cell mass.
\item Day 5: The morula continues to develop to form the blastocyst. The blastocyst has a distinct BlastoCoel (BC), and the internal cells begin to differentiate to form two main parts: the Inner Cell Mass (ICM), which will develop into the fetus; The TrophEctoderm (TE) develops into the placenta. The Zona Pellucida (ZP) of the blastocyst begins to thin and is ready for hatching and implantation.
\end{itemize}
\indent In clinical practice, cleavage embryos (day 3) or blastocyst embryos (day 5) are usually used for transfer. Cleavage embryos are used because some patients have a low number of embryos and are concerned about losing embryos to culture at the blastocyst stage. In addition, due to limited culture conditions in some laboratories, the success rate of culture of cleavage embryos is higher than that of blastocyst. Blastocyst embryos are selected because after an additional 2-3 days of culture, embryo quality is naturally screened and inferior embryos are eliminated. In addition, if PGT genetic screening is needed, blastocyst embryos are more suitable because there are more cells in the embryo to sample.\\
\indent During the IVF-ET process, there are two key human decision-making points, step 4 and step 5 in the Fig. \ref{step}(b) and (c), which give rise to the two tasks of embryo grading and pregnancy prediction. For the task of embryo grading, embryologists need to select the embryo with the most developmental potential as the subsequent transfer object according to the characteristics of embryo development, division rate, cell symmetry and so on. For the task of pregnancy prediction, reproductive doctors combine the embryo grading results with parental fertility indicators to comprehensively determine the embryo transfer scheme and optimize the pregnancy success rate. Therefore, embryo grading and pregnancy prediction have become two core tasks in the clinical practice of assisted reproduction and are highly dependent on professional experience. The application of automated decision-making systems in embryo grading and pregnancy prediction has important clinical value and development potential.

\subsection{The second generation——ICSI}
ICSI was first successfully applied in clinical practice in 1992 by a research team from the Free University of Brussels in Belgium. This breakthrough marked an important step in the field of ART. ICSI is mainly used for the treatment of infertility caused by male factors, especially severe oligospermia, asthenospermia, and azoospermia. Although the overall procedure of ICSI is similar to that of conventional IVF-ET, there are significant differences in sperm acquisition and fertilization stages \cite{22}. In the sperm acquisition stage, ICSI has relatively low requirements for sperm quality, and sperm can be obtained by testicular or epididymal puncture even for azoospermia patients. During the fertilization stage, ICSI uses a micromanipulative technique to inject a single sperm directly into the cytoplasm of the egg cell, thereby bypassing the natural sperm penetration through the zona pellucidum and cell membrane of the egg cell. This technique has significantly improved the fertilization success rate of male infertility patients and brought hope to many families.

\subsection{The third generation——PGT}
PGT is a revolutionary technology in the field of ART. PGT uses genetic analysis of early embryos cultured in vitro to screen embryos without specific genetic abnormalities for transfer, aiming to reduce the abortion rate, improve the success rate of pregnancy, and effectively block the transmission of single-gene genetic diseases or chromosomal abnormalities to offspring \cite{23}. PGT is generally performed at the blastocyst stage, usually on day 5-6 after fertilization, and 5 to 10 TE cells are taken from the blastocyst for testing, without involving the ICM that will develop into the fetus in the future. Blastocyst stage biopsy can not only ensure the embryo development potential, but also provide more adequate genetic material for analysis. Through PGT technology, precise genetic screening and disease prevention can be achieved before embryo implantation, which greatly improves the overall success rate of ART.

% \section{Section title of first appendix}\label{secA1}

% An appendix contains supplementary information that is not an essential part of the text itself but which may be helpful in providing a more comprehensive understanding of the research problem or it is information that is too cumbersome to be included in the body of the paper.




\end{appendices}


\bibliography{sn-bibliography}% common bib file

\section*{Author Biographies}

\noindent
\begin{minipage}[c]{0.22\textwidth}
  \includegraphics[width=\linewidth]{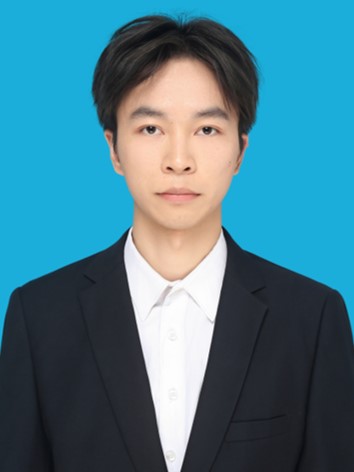}
\end{minipage}%
\hfill
\begin{minipage}[c]{0.75\textwidth}
  \textbf{Xueqiang Ouyang} is an Assistant Engineer at the Information Department, The People's Hospital of Baoan Shenzhen, China. He received his Bachelor's and Master's degrees from the South China University of Technology in 2022 and 2025. His main research is Artificial Intelligence in Medicine.
\end{minipage}

\vspace{1.5em}

\noindent
\begin{minipage}[c]{0.22\textwidth}
  \includegraphics[width=\linewidth]{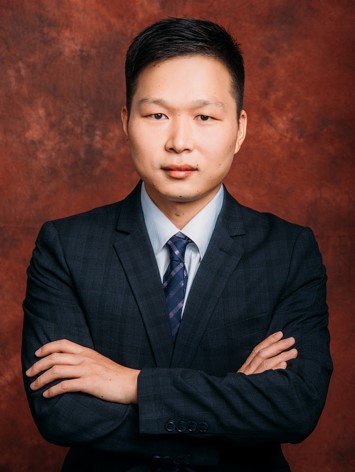}
\end{minipage}%
\hfill
\begin{minipage}[c]{0.75\textwidth}
  \textbf{Jia Wei} is an Associate Professor at the School of Computer Science and Engineering, South China University of Technology, China. He received his Bachelor's and Master's degrees in Computer Science from the Harbin Institute of Technology in 2003 and 2006, and his Ph.D. from the South China University of Technology in 2009. His main research interests include Machine Learning, Deep Learning, Artificial Intelligence in Medicine.
\end{minipage}

\end{document}